\documentclass[runningheads]{llncs}
\usepackage{graphicx}

\usepackage{tikz}
\usepackage{comment}
\usepackage{amsmath,amssymb} 
\usepackage{color}
\usepackage{epsfig}
\usepackage{multirow}
\usepackage{subcaption}
\usepackage{bbm}
\usepackage{array}
\usepackage{booktabs}
\usepackage{kotex}
\usepackage{wrapfig}
\usepackage{enumitem}
\newcolumntype{P}[1]{>{\centering\arraybackslash}p{#1}}

\usepackage[accsupp]{axessibility}  

\begin{document}
\pagestyle{headings}
\mainmatter
\def\ECCVSubNumber{3958}  

\title{Unsupervised Domain Adaptation for One-stage Object Detector using Offsets to Bounding Box}


\titlerunning{UDA for One-stage Object Detector using Offsets to Bounding Box}
%
\author{Jayeon Yoo\inst{1} \and
Inseop Chung\inst{1} \and
Nojun Kwak\inst{1}}
\authorrunning{J. Yoo et al.}
%
\institute{Seoul National University \\
\email{\{jayeon.yoo,jis3613,nojunk\}@snu.ac.kr}}
\maketitle

\begin{abstract}
Most existing domain adaptive object detection methods exploit adversarial feature alignment to adapt the model to a new domain. Recent advances in adversarial feature alignment strives to reduce the negative effect of alignment, or negative transfer, that occurs because the distribution of features varies depending on the category of objects. However, by analyzing the features of the anchor-free one-stage detector, in this paper, we find that negative transfer may occur because the feature distribution varies depending on the regression value for the offset to the bounding box as well as the category. To obtain domain invariance by addressing this issue, we align the feature conditioned on the offset value,
considering the modality of the feature distribution. With a very simple and effective conditioning method, we propose OADA (Offset-Aware Domain Adaptive object detector) that 
achieves state-of-the-art performances in various experimental settings. In addition, by analyzing through singular value decomposition, we find that our model enhances both discriminability and transferability.
\end{abstract}

\section{Introduction}
Deep-learning-based object detection has shown successful results by learning from a large amount of labeled data. However, if the distribution of test data is significantly different from that of training data, the model performance is severely impaired.
In practice, this performance degradation can be very fatal because the domains in which the object detection model should operate can be very diverse. 
To address this problem, the most effective way is to re-train the model with a lot of data from the new environment whenever the environment changes.
However, obtaining a large amount of labeled data is a very expensive process, especially in object detection task which requires annotating the bounding boxes and the classes of objects in an image. Unsupervised Domain Adaptation (UDA) provides an efficient solution to this domain-shift problem in a way that it adapts the model to a new environment by training with unlabeled datasets from the new environment (target domain) as well as rich datasets from the original environment (source domain). 
Based on the theoretical analysis of \cite{ganin2015dann},
aligning the feature distribution of the source and the target domain in an adversarial manner is one of the most effective ways in various tasks such as classification \cite{ganin2015dann,xu2020mixup,caron2018clustering,na2021fixbi} and segmentation \cite{pa2020segmentation,yang2020fda,chung2022interpixel,chung2022cosine}. 
A seminal work \cite{chen2018domain} is the first to deal with \textit{Domain Adaptive Object Detection} (DAOD) aligning backbone features via an adversarial method, and many follow-up studies have continued in this line of research. 
Unlike classification and segmentation tasks classifying an image and each pixel as one category, object detection is a task of classifying the categories and regressing the bounding box of each foreground object.
Focusing on this difference, many studies further align local features \cite{saito2019strongweak} or instance-level features corresponding to the foreground rather than the background \cite{hsu2020epm,xu2020cateregul}. 

While one-stage detectors such as FCOS~\cite{tian2019fcos} and YOLO~\cite{redmon2016yolo} are more advantageous for real-world environments because of its efficient structures and high inference speed, most DAOD studies \cite{chen2018domain,saito2019strongweak,hsu2019progressive,deng2021unbiased,chen2020harmonizing,vs2021mega,zhao2020adaptive} have been conducted on two-stage detectors, such as Faster R-CNN~\cite{ren2016faster_rcnn}. They use proposals generated by Region Proposal Network (RPN) to obtain instance-level features corresponding to the objects, making it difficult to extend straightforwardly to one-stage detectors that do not rely on RPN. Recently, several DAOD methods specialized in one-stage detector have been proposed \cite{hsu2020epm,munir2021ssal}.
They prevent negative transfer that can occur when a feature is indiscriminately aligned by focusing on a foreground object or further aligning a feature according to the category of the object. However, for FCOS, an one-stage detector that estimates offsets from each point of the feature map to the four sides of a bounding box, the features differ in distribution not only by categories but also by offsets. Accordingly, existing feature alignment may not be sufficient to prevent negative transfer.

\begin{figure*}[t]
\centering
  \begin{subfigure}[t]{4cm}
    \includegraphics[width=.9\linewidth]{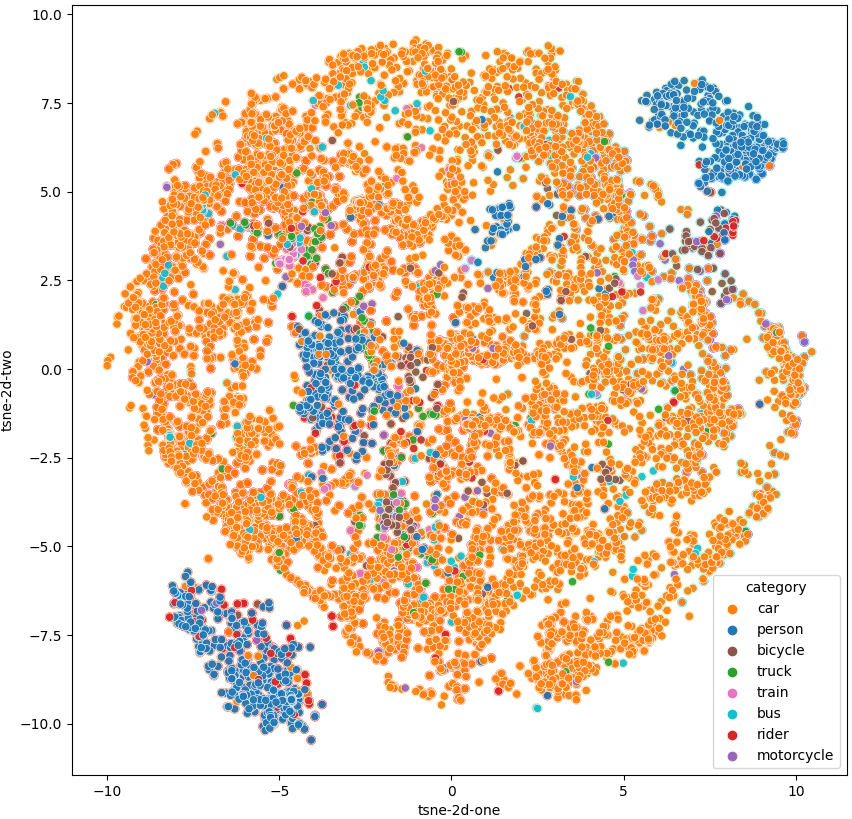}
    \caption[width=\linewidth]{According to the classes. Orange and blue indicate \textit{car} and \textit{person} respectively.}
    \label{fig:TSNE-CLS}
  \end{subfigure}
  \centering
  \begin{subfigure}[t]{4cm}
    \captionsetup{margin=0.2cm}
    \includegraphics[width=.9\linewidth]{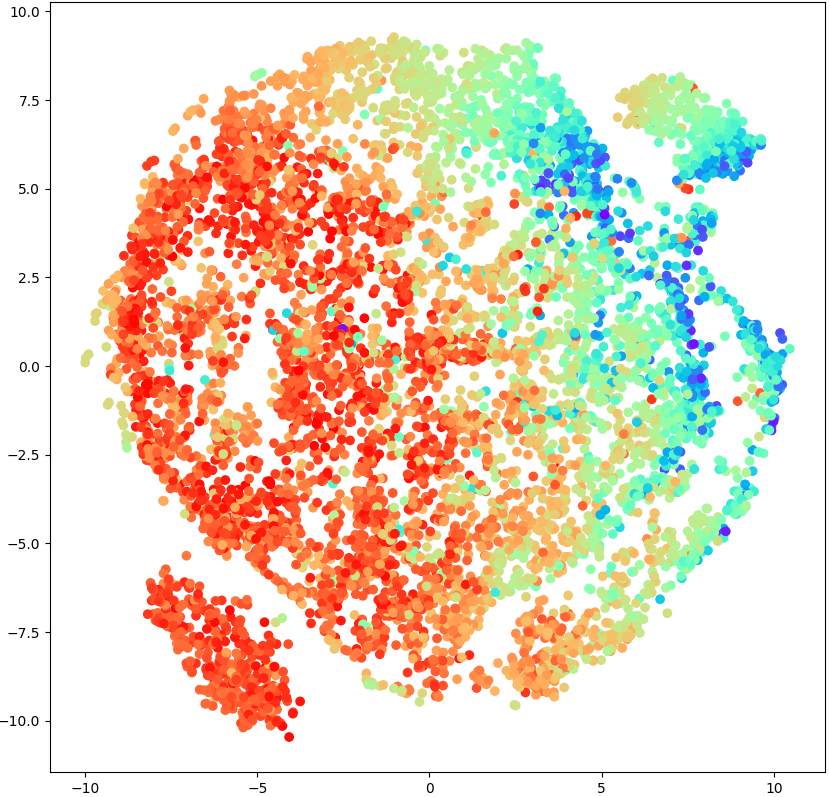}
    \caption{According to the offset to the top side of the GT bounding box ($t$).}
    \label{fig:TSNE-REG-TOP}
  \end{subfigure}
  \centering
  \begin{subfigure}[t]{4cm}
    \captionsetup{margin=0.2cm}
    \includegraphics[width=.9\linewidth]{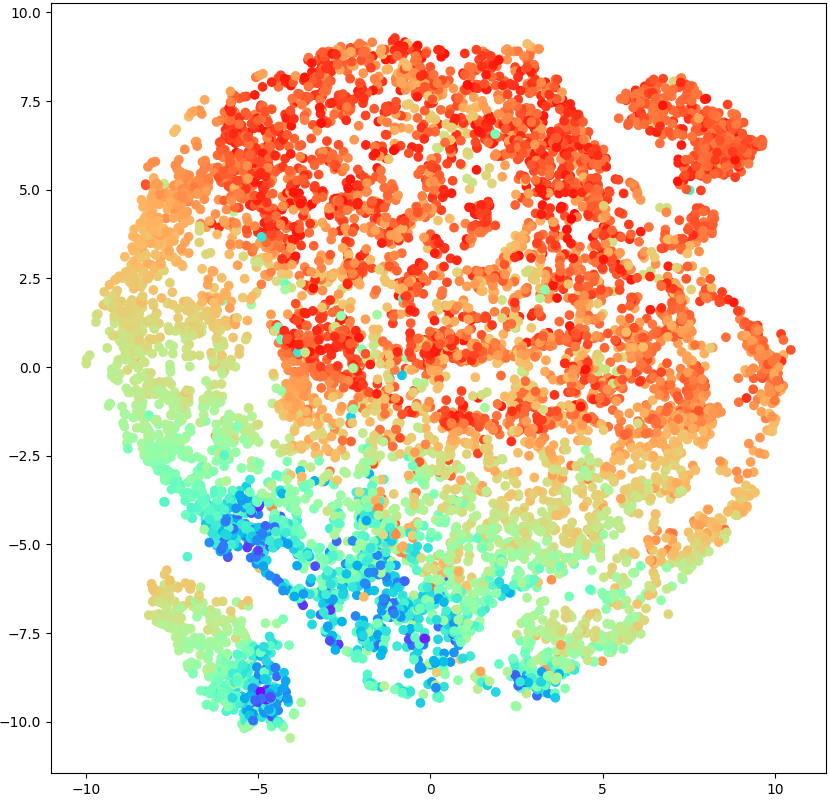}
    \caption{According to the offset to the bottom side of the GT bounding box ($b$).}
    \label{fig:TSNE-REG-BOTTOM}
  \end{subfigure}
  \caption{
  TSNE visualization of the source domain backbone features of a detector that is trained only on the source domain. Different colors in (a) refer to different classes while in (b) and (c), the colors represent the distance from a location of a feature with high classification confidence to the top and bottom of the ground-truth bounding box respectively. The redder the color, the greater the distance.
  }
  \label{fig:TSNE}
\end{figure*}

Fig. \ref{fig:TSNE} shows the TSNE of features corresponding to foreground objects obtained from the backbone of FCOS trained on the Cityscape dataset \cite{cordts2016cityscapes} consisting of 8 classes. In Fig. \ref{fig:TSNE-CLS}, different colors refer to different classes. Since instances of \textit{Car} (orange) and \textit{Person} (blue) are dominant, the difference in feature distribution between the \textit{Car} and the \textit{Person} is clearly visible. 
This phenomenon fits well with the intention of \cite{chen2021i3net,vs2021mega,zhao2020dual,xu2020graph,zheng2020coarse-fine} which align the feature distribution of the source and the target domain in a class-wise manner. Fig. \ref{fig:TSNE-REG-TOP} and \ref{fig:TSNE-REG-BOTTOM} show the feature distribution in another perspective, the distance to the boundary of the GT bounding box. In Fig. \ref{fig:TSNE-REG-TOP}, color codes are used to measure the offset, the distance from the feature point to the top side of the bounding box, in the log scale: the redder, the larger the offset is, while the bluer, the smaller.
It shows that the distribution of the backbone features varies markedly with offsets.
Comparing Fig. \ref{fig:TSNE-CLS} and \ref{fig:TSNE-REG-TOP}, even features in the same category have varying distribution depending on their offset.
Since object detection requires bounding box regression as well as classification, especially in case of FCOS which predicts the offsets of \textit{(left, top, right, bottom)} to the four sides of bounding boxes,
the backbone features are not only clustered by categories but also distributed according to the offsets. Paying attention to this analysis, we conditionally align features of the source and the target domains according to their offsets.

Therefore, in this paper, we propose an Offset-Aware Domain Adaptive object detection method (OADA) that aligns the features of the source and the target domain conditional to the offset values to suppress the negative transfer in an anchor-free one-stage detector such as FCOS. Specifically, to align instance-level features and obtain reliable offset values, we use label information for the source domain and classification confidence for the target domain. And then, we convert continuous offsets into categorical probability vectors and get offset-aware features by outer-producting that probability vectors and backbone features.
We prevent negative transfer that may occur while aligning the features to have the same marginal distribution by making the offset-aware features domain-invariant using a domain discriminator. 
Essentially, this is equivalent to intentionally forming a discriminator embedding space that is roughly partitioned by the offset. 
As a result, we can efficiently align features with a single strong discriminator, opening up new possibilities for offset-aware feature alignment in a very simple yet effective manner. Our contributions can be summarized as follows:
\begin{itemize}
    \item We present a domain adaptation method which is specialized for anchor-free one-stage detector by analyzing the characteristics of it.
    \item We prevent negative transfer when aligning instance-level features in domain adaptive object detection by making domain-invariant offset-aware features in a highly efficient manner.
    \item{We find that our proposed method enhances both discriminability and transferability by analyzing through singular value decomposition.}
    \item{We show the effectiveness of our proposed method (OADA) through extensive experiments on three widely used domain adaptation benchmarks, Cityscapes $\rightarrow$ Foggy Cityscapes and Sim10k, KITTI $\rightarrow$ Cityscapes and it achieves state-of-the-art performance in all benchmarks.}
\end{itemize}
\bigskip

\section{Related Works}
\label{subsec:related_works}

\subsection{Object Detection}
Deep-learning-based object detection can be categorized into anchor-based and anchor-free methods. Anchor-based detectors define various sizes and ratios of anchors in advance and utilize them to match the output of the detector with the ground-truth. On the other hand, anchor-free detectors do not utilize any anchors but rather directly localize objects employing fully convolutional layers. Moreover, depending on whether region proposal network (RPN) is used or not, object detectors can also be classified into two-stage and one-stage detectors.
Faster R-CNN \cite{ren2016faster_rcnn} is a representative anchor-based two-stage detector while SSD \cite{liu2016ssd} and YOLO \cite{redmon2016yolo} are anchor-based one-stage detectors. There are some renowned anchor-free one-stage detectors as well. Cornernet \cite{law2018cornernet} and Centernet \cite{duan2019centernet} localize objects by predicting the keypoints or the center of an object while FCOS \cite{tian2019fcos} directly computes the offset from each location on the feature map to the ground-truth bounding box. Most works of DAOD have been conducted on Faster R-CNN, a two-stage detector and relatively few works have been done using an anchor-free one-stage detector. There are several works \cite{chen2021i3net,rodriguez2019style,kim2019selftraining} that have been conducted on SSD, a representative one-stage anchor-based detector and only \cite{hsu2020epm,munir2021ssal} carried out domain adaptation using FCOS, an anchor-free one-stage detector. Our work focuses on boosting the domain adaptation performance on FCOS \cite{tian2019fcos} leveraging its anchor-free architecture and fast speed.

\subsection{UDA for Object Detection}
There are three main approaches of UDA for object detection tasks: adversarial alignment, image translation, and self-training. Image-translation-based methods translate the source domain images into another domain using a generative model \cite{kim2019diversify,deng2021unbiased,shan2018pixel,chen2020harmonizing,rodriguez2019style,hsu2019progressive} to adapt to the target domain. 
Self-training-based methods \cite{roy2019automatic,khodabandeh2019robust,deng2021unbiased,kim2019selftraining,munir2021ssal} generate pseudo-labels for the target domain images with the model pre-trained on the source domain and re-train the model with the pseudo labels. 
For adversarial alignment methods, \cite{chen2018domain} is a seminal work that aligns the feature distribution of the source and the target domain using a domain discriminator based on the Faster R-CNN \cite{ren2016faster_rcnn}. Since then, there have been studies to align feature distribution at multiple levels \cite{saito2019strongweak,he2019multi}, studies focusing on the importance of local features in detection, and studies to align instance-level features that may correspond to objects \cite{xu2020exploring,zhao2020adaptive,hsu2020epm}. Recently, there have been attempts \cite{chen2021i3net,vs2021mega,zhao2020dual,xu2020graph,zheng2020coarse-fine} to align the instance-level features in a class-wise manner, focusing on the fact that the distribution of instance-level features is clustered by class. Based on our observation that the feature distribution varies depending on the offset values in FCOS \cite{tian2019fcos} and the detection task requires not only classification but also regression, we propose an adversarial alignment scheme with state-of-the-art performances in various experimental settings by aligning the features in an offset-aware manner.

\section{Method}
\begin{figure*}[t]
\centering
\includegraphics[width=.8\linewidth]{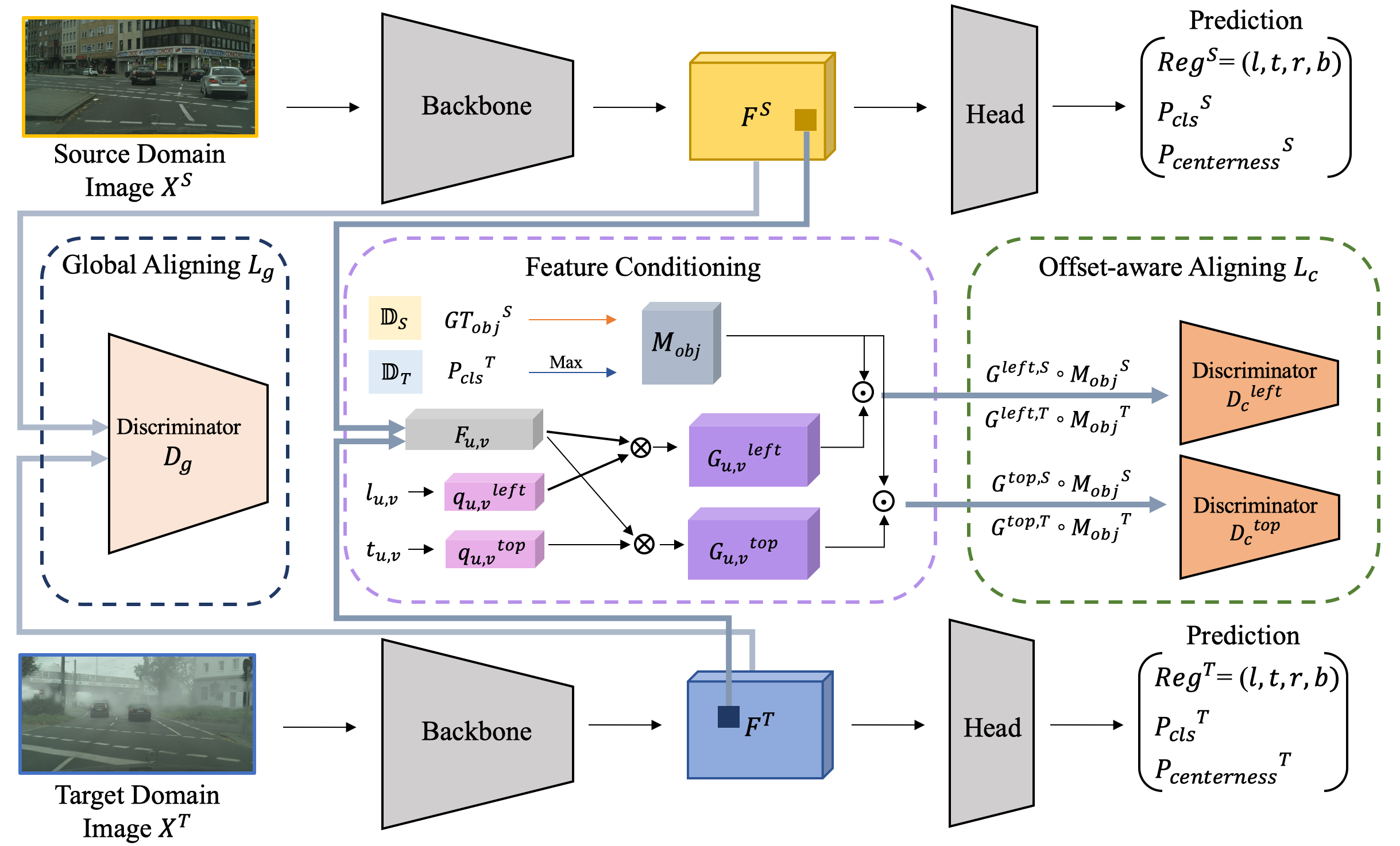}
\caption{The overall structure of our framework. $F^S$ and $F^T$ are the $lv$-th level features ($lv \in \{3,..,7\}$, $lv$ is omitted for simple notation) of the source image $X^S$ and the target image $X^T$, respectively. The overall feature maps $F^S$ and $F^T$ are aligned by the global domain discriminator, $D_{g}$. To generate a mask $M_{obj}$ corresponding to the objects, the GT labels are used for the source domain and maximum values of class confidence higher than threshold is used for the target domain. The GT offsets of the source domain and the predicted offsets of the target domain are converted into probability vector $q_{u,v}$, and they produce $G_{u,v}$ via the outer-product with $F_{u,v}$.
We use $D_{c}^{left}$ and $D_{c}^{top}$ to align the conditioned features $G_{u,v}^{l}$ and $G_{u,v}^{t}$ according to the left and top offsets, respectively.}
\label{fig:overview}
\end{figure*}

In this section, we describe our method shown in Fig. \ref{fig:overview}, which aligns instance-level features between the two domains in an offset-aware manner in detail. Since our method investigates domain adaptation of a representative anchor-free one-stage detector, FCOS~\cite{tian2019fcos}, a brief introduction about it is given in  Sec. \ref{subsec:preliminary}.

\subsection{Preliminary: FCOS}
\label{subsec:preliminary}

\begin{figure}[t]
\centering
\begin{minipage}{.55\textwidth}
  \centering
  \includegraphics[width=.8\linewidth,height=70pt]{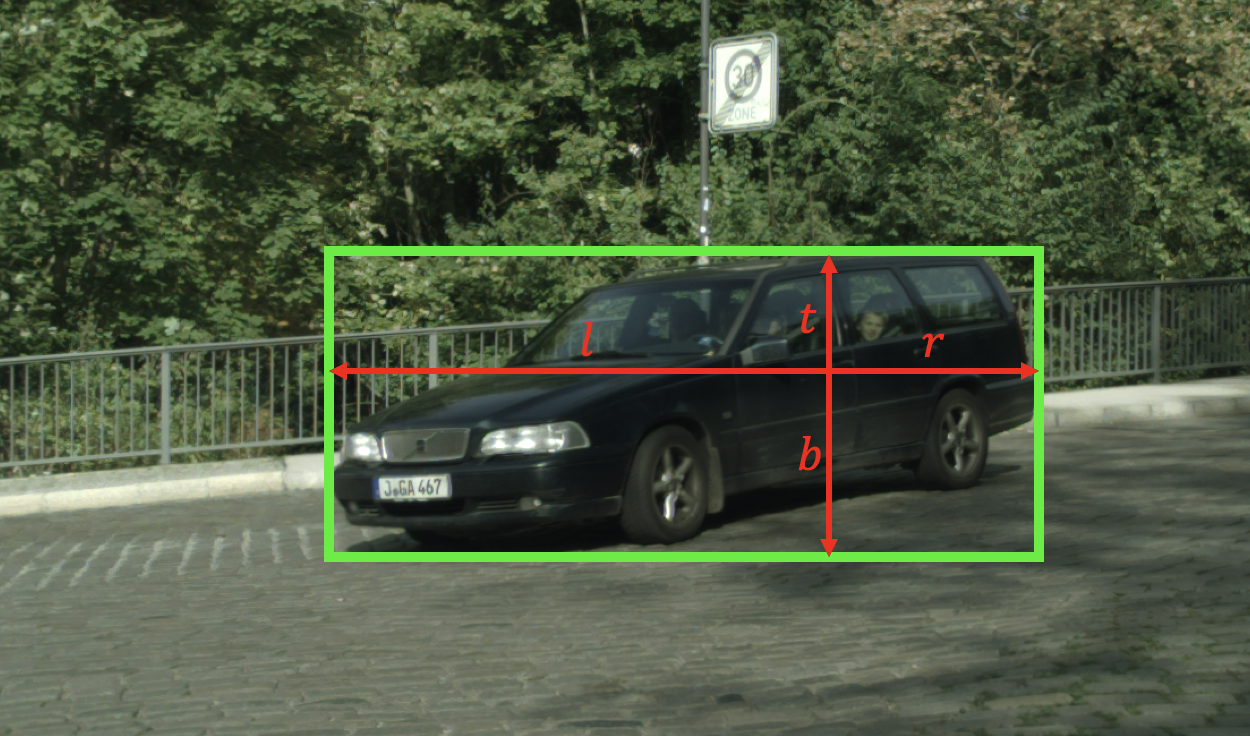}
  \captionof{figure}{FCOS predicts left, top, right, and bottom $(l, t, r, b)$ distances to the bounding box from each location of the feature map.
}
  \label{fig:FCOS}
\end{minipage}%
\begin{minipage}{.43\textwidth}
  \centering
  \captionsetup{margin=0.2cm}
  \includegraphics[width=0.8\linewidth,height=70pt]{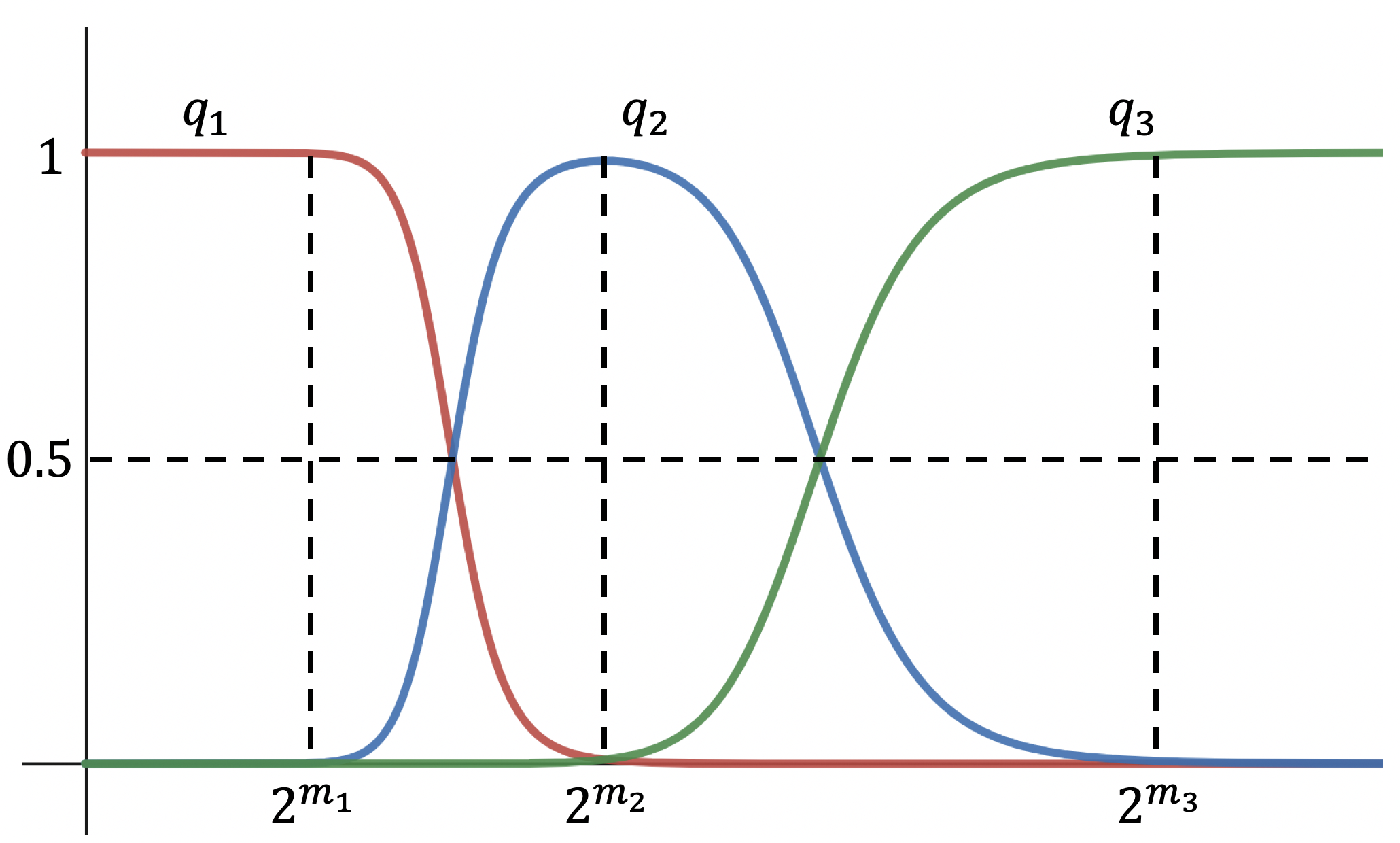}
  \captionof{figure}{The probability distribution of the regression value belonging to each of three bins.}
  \label{fig:Regression_Probability}
\end{minipage}
\end{figure}

FCOS~\cite{tian2019fcos} is a representative one-stage detector that predicts object categories and bounding boxes densely in feature maps without a RPN. FCOS uses five levels of feature maps ($F_3 \sim  F_7$) produced from the backbone network following 
FPN~\cite{lin2017fpn} to detect various sizes of objects. At each location of the feature map, it predicts the corresponding object category, the centerness indicating how central the current location is to the object, and the distances from the current location to the left, top, right, and bottom $(l, t, r, b)$ of the nearest ground-truth bounding box as shown in Fig. \ref{fig:FCOS}. With a design that five levels of feature maps have different resolutions decreasing by a factor of $1/2$, the maximum distance responsible for each feature level is set to ($64, 128, 256, 512,\infty$). 

FCOS differs from conventional anchor-based or RPN-based object detectors which regress four values to correct anchors or proposals.
We observe that the features are distributed according not only to object categories but also to offsets, the distances to the nearest bounding boxes, and it is more pronounced in FCOS due to its characteristics of predicting $(l, t, r, b)$ directly using the features. Focusing on this observation, our proposed method tries to align the features of the source and the target domains by conditioning the features with the offsets.

\subsection{Problem Formulation}
\label{subsec:ProblemFormulation}
Consider the setting where we adapt the object detector to the target domain using labeled source domain data $\mathbb{D}_S$ and unlabeled target domain data $\mathbb{D}_T$ which share the same label space consisting of $C$ classes. When training a detector, we only have access to $\mathbb{D}_S = {(x_i^S, y_i^S, b_i^S)}_{i=1}^{N_S}$ and $\mathbb{D}_T = {(x_i^T)}_{i=1}^{N_T}$, where $x_i$ is the input image, and $y_i \in [C]^{k\times 1}$ and $b_i \in \mathbb{R}^{k\times 4}$ are the object categories and the bounding box coordinates of all the $k$ objects existing in $x_i$. $N_S$ and $N_T$ are the numbers of samples in $\mathbb{D}_S$ and $\mathbb{D}_T$.

\subsection{Global Alignment}
\label{subsec:global_alignment}
To ensure that the object detector works well on the target domain, we align the features of both the source and the target domain to have a marginally similar distribution through an adversarial aligning method using a global domain discriminator $D_{g}$. Let $F^S$ and $F^T$ be the feature maps obtained by feeding the source domain image $x^S$ and the target domain image $x^T$ to the backbone, respectively. When the spatial size of a feature map $F$ is $H\times W$, $D_{g}$ is trained to classify the domain of the feature map $F$ pixel-wisely by the binary cross entropy loss as in (\ref{eq:loss-global}). The label of the source domain is 1, while that of the target is 0. 
\begin{equation}
    \begin{gathered}
        \mathcal{L}_{g}(x^{S}, x^{T})= - \sum_{u=1}^H \sum_{v=1}^{W} \log(D_{g}(F^S_{u,v}))+\log(1-D_{g}(F^T_{u,v})) .
    \end{gathered}
\label{eq:loss-global}
\end{equation}
Using the gradient reversal layer (GRL) proposed in \cite{ganin2015dann}, the backbone is adversarially trained to prevent the domain discriminator from correctly distinguishing the source and the target domains, thereby generating domain-invariant features.

\subsection{Generating Conditional Features to Offset Values}
\label{subsec:generating_conditional_features}
In order to align features in an offset-aware manner, the features and the corresponding offsets can be concatenated and inputted to the domain discriminator. 
However, simply concatenating them is not enough to fully utilize the correlation between features and offset values. Inspired by \cite{long2018conditional}, which considers the correlation between features and categorical predictions, our method does not simply concatenate but outer-product features and offset values. Unlike classification which uses a categorical vector, an offset value is a continuous real value. Hence, the product of the feature and the offset value only has the effect of scaling the feature.
To effectively condition the feature according to the corresponding offset values, each offset value of $(l, t, r, b)$ is converted into a probability vector corresponding to $N_{bin}$ bins using (\ref{eq:reg-bin}). In the equation, $z_{u,v}^n$ refers to the offset value for $n \in \{l,t,r,b\}$ at the location $(u,v)$ in the feature map.
To convert the offset $z_{u,v}$ to an $N_{bin}$-dimensional probability vector, we calculate the probability using the distance of the log of the offset value to a predefined value $m_i$ for the $i$-th bin. Note that we have $N_{bin}$ bins and $m_1<m_2<\cdots<m_{N_{bin}}$. 
Assuming that the probability of $z_{u,v}$ belonging to the $i$-th bin is proportional to a normal distribution with its mean $m_i$ and a shared variance $\sigma^2$, it becomes as follows:
\begin{equation}
    \begin{gathered}
        q_{i}(z_{u,v}^n) = \frac{\exp(-\frac{(\log_2(z_{u,v}^n)-m^n_i)^2}{2  \sigma^2 / \tau})}{\sum_{j=1}^{N_{bin}} \exp(-\frac{(\log_2(z_{u,v}^n)-m_j^n)^2}{2  \sigma^2 / \tau})}, i \in [N_{bin}], n \in \{l, t, r, b\}.
    \end{gathered}
\label{eq:reg-bin}
\end{equation}

Here, $\tau$ is a temperature value to make the distribution smooth, and in all of our experiments, both $\sigma$ and $\tau$ are set to 0.1. In all of our main experiments, $N_{bin}$ is set to 3 for each feature level, and $m_i$ for each bin is set to $(m_1,m_2,m_3)=(lv-\frac{1}{2},lv+\frac{1}{2},lv+\frac{3}{2})$ for $lv$-th level ($lv \in \{3,...,7\}$) to satisfy $\frac{m_1+m_2}{2}=lv$ and $\frac{m_2+m_3}{2}=lv+1$ because each feature level is responsible for a different object scale. Fig. \ref{fig:Regression_Probability} shows the probability of $z_{u,v}$ belonging to each bin when there are three bins. The obtained probability vector ${q} \triangleq [q_1, \cdots, q_{N_{bin}}]^T \in \mathbb{R}_+^{N_{bin}}$ still maintains the relative distance relationship of the real offset value as $D_{KL}({q}(a) || {q}(b)) < D_{KL}({q}(a) || {q}(c))\  \text{if}\ \ a < b < c$. 

However, the probability vector is uniformly initialized for all ${N_{bin}}$ bins since the regressed offsets may not be accurate at the beginning of training. During the first $I$ warm-up iterations, we gradually increase the rate of using the probability vector ${q}_{u,v}$ as iteration ($iter$) progresses utilizing the alpha-blending as follows: 
\begin{equation}
    \begin{gathered}
        \tilde{{q}}_{u,v} = ((1-\alpha) {q}_{u,v} + \alpha\frac{1}{N_{bin}} \mathbbm{1}), \quad \text{where} \quad \alpha = \max(1-\frac{iter}{I}, \alpha_0).
    \end{gathered}
\label{eq:alpha-blending}
\end{equation}

Here, $\alpha_0$ is the constant value between 0 and 1 that smoothes the probability vector $\tilde{q}$, and the closer it is to 1, the more uniform $\tilde{q}$ becomes. We show the effects of $\alpha_0$ in Sec.\ref{subsec:ablation}. In all the experiments, $I$ is set to 6k, the half of the first learning rate decay point. $\mathbbm{1}$ is a vector consisting of only ones.

Using $\tilde{{q}}_{u,v}$, we can obtain the features which are conditional to the offset values by outer-producting them as follows:
\begin{equation}
    \begin{gathered}
        g_{u,v} = f_{u,v}\otimes \tilde{q}_{u,v}.
    \end{gathered}
\label{eq:reg-conditioning}
\end{equation}

Here, $f_{u,v}\in \mathbb{R}^{D}$ is a feature vector located at location $(u,v)$ in the feature map $F \in \mathbb{R}^{D \times HW}$ and $\tilde{q}_{u,v} \in \mathbb{R} ^{N_{bin}}_+$ is the probability vector of the corresponding offset value obtained from (\ref{eq:alpha-blending}).  
By outer-producting $f_{u,v}$ and $\tilde{q}_{u,v}$, we can get a new feature, $g_{u,v} \in \mathbb{R}^{D \times N_{bin}}$ conditioned on the offsets. 
By flattening $g_{u,v}$, the conditioned feature map $G \in \mathbb{R}^{(D \times N_{bin}) \times HW}$ with the same spatial resolution as $F$ is obtained, which is fed into the domain discriminator $D_c$.

Outer product is effective in conditioning because it considers the correlation between features and offsets without loss of information, hence it enables features to have different characteristics depending on offset values. Consider a case where $N_{bin}=3$ and $(m_1, m_2, m_3)=(3.5, 4.5, 5.5)$ for $F_4$, and conditionally align to $t$, the top offsets.
Suppose that feature vector $f_{u_1,v_1} \in \mathbb{R}^D$ located at $(u_1,v_1)$ have a small top offset prediction, i.e. $t=13$ and $\log_2 t = 3.7$, resulting in the probability vector $q_{u_1,v_1}=(0.98,0.02,0.0)$ via (\ref{eq:reg-bin}).
Conditioned feature $g_{u_1,v_1}$ obtained by outer-producting $f_{u_1, v_1}$ and $q_{u_1, v_1}$ is a $3\times D$ matrix. 
The first row of $g_{u_1,v_1}$ would be similar to the original feature $f_{u_1,v_1}$, but the elements in the other rows would be almost zero. On the other hand, $g_{u_2,v_2}$ obtained by e.g. $q_{u_2,v_2}=(0,0.01,0.99)$ with a large top offset would have the original feature $f_{u_2,v_2}$ in the third row but have almost zero elements in another rows.
Therefore, a domain discriminator is trained to classify the domain of the features in different subspaces according to the offsets. As a result, the backbone will generate features that are domain invariant conditioned on offsets to fool the discriminator.

\subsection{How to get a confident offset value?}
\label{subsec:WheretoAlign}

\begin{figure*}[t]
\centering
  \begin{subfigure}[t]{5cm}
    \includegraphics[width=\linewidth,height=80pt]{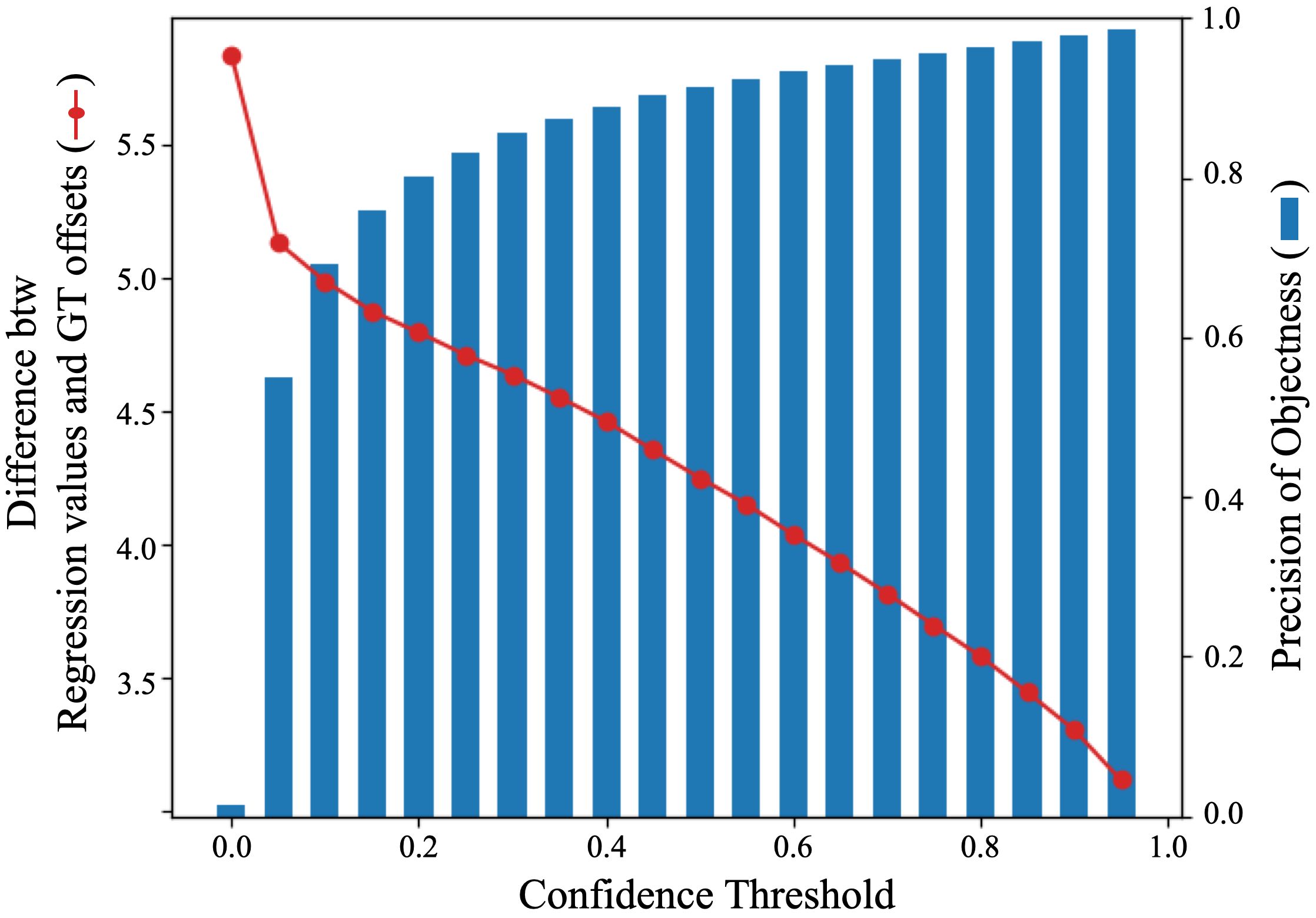}
    \caption[width=\linewidth]{Cityscapes$\rightarrow$Foggy Cityscapes.}
    \label{fig:Conf-CS}
  \end{subfigure}
  \centering
  \begin{subfigure}[t]{5cm}
    \captionsetup{margin=0.2cm}
    \includegraphics[width=\linewidth,height=80pt]{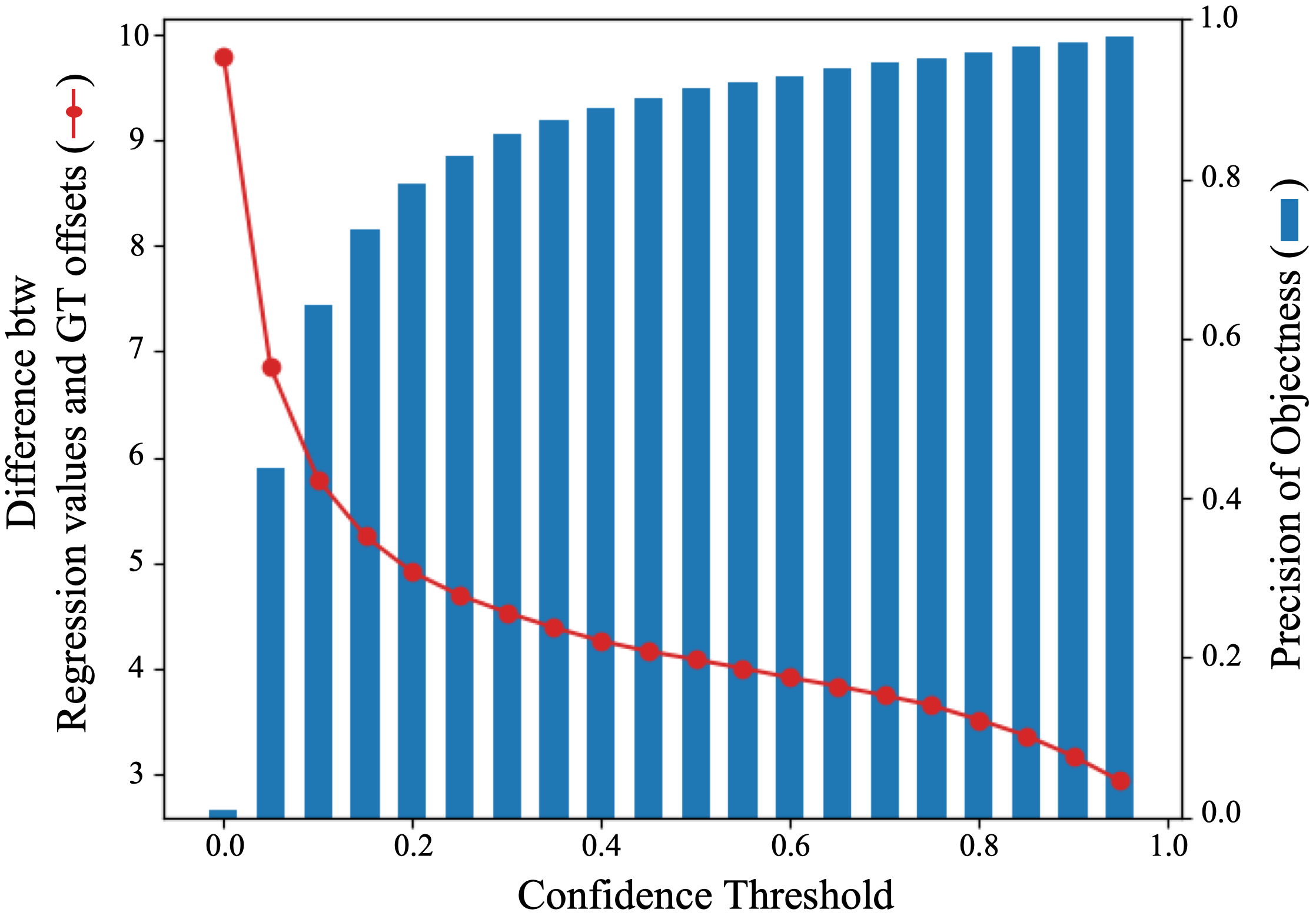}
    \caption{Sim10k$\rightarrow$Cityscapes.}
    \label{fig:Conf-Sim10k}
  \end{subfigure}
  \caption{The difference between the predicted regression value and the GT value (Red line) and the precision (Blue bar) according to the confidence threshold $\rho$.}
  \label{fig:Conf-calibration}
\end{figure*}

To generate a feature conditioned on the offsets, we need to know which location in the feature map corresponds to the object and the accurate offset value at that location. For the labeled source domain, we can easily obtain the ground-truth value of which location corresponds to the object in each feature map and the offsets. Since FCOS calculates the object mask corresponding to each category in the feature map and the distance from each location to the GT bounding box, we can utilize this mask and the ground-truth offset values, $(l^*, t^*, r^*, b^*)$. On the other hand, in the unlabeled target domain, we should inevitably use predicted values. Although classification confidence is the probability of predicting the category of an object rather than a regression, by using it, we can simply and effectively select instance-level features with high objectness and obtain reliable regression values for those features. Fig.~\ref{fig:Conf-calibration} is the analysis of features obtained by feeding a target domain image into the globally aligned model of Sec. \ref{subsec:global_alignment}. 
The blue bar represents the ratio corresponding to the actual objects among the features having a confidence value higher than the threshold value of the x-axis, and the red line represents the average of the difference between the regression values and the GT offsets of the features. 
The higher the confidence threshold, the higher the probability of the feature's location belonging to an actual object, and the closer the regression value is to the GT offset value.
In $p_{cls}^{u,v} \in \mathbb{R}_+^{C}$, which is the category classification probability at the $(u,v)$-th location of the feature map, $\max_{c \in [C]}(p_{cls}^{u,v})$ (the maximum probability among all classes) can be viewed as objectness which is the probability that the location corresponds to an object. Therefore, we generate a mask where $\max_{c \in [C]}(p_{cls}^{u,v})$ is higher than a threshold $\rho$, and we weight the activated part with that max probability value as follows:
\begin{equation}
    \begin{gathered}
        M_{obj}^T = \mathbbm{I}_{\max_{c}(p_{cls}) > \rho} \odot \max_{c}(p_{cls}). 
    \end{gathered}
\label{eq:mask}
\end{equation}
Finally, we align the distribution of features $G$ of the source and the target which is conditioned on the offset by minimizing the following adversarial loss:
\begin{equation}
\begin{split}
\mathcal{L}_{c}^{n}(x^S, x^T) &= -\sum_{u,v}^{H,W} \log(D_{c}^{n}(\hat{G}^{n,S}_{u,v}))+\log(1-D_{c}^{n}(\hat{G}^{n,T}_{u,v})) \\
&\hat{G}^{n,d} = G^{n,d} \odot M_{obj}^{d},  \quad  d \in \{S,T\}.
\end{split}
\label{eq:cond_loss}
\end{equation}
Here, $\mathcal{L}_{c}^{n}$ is the adversarial loss conditional to the offset value $n \in \{l, t\}$ using the discriminator $D_c^n$. Since the correlation between the regression values for the left $l$ and the right $r$ and between the top $t$ and the bottom $b$ are strong, conditioning is performed only for the left and the top. $d$ represents whether the domain is the source or the target. $\mathbbm{I}$ is the indicator function and $\odot$ is the elementwise multiplication.
Through this, the instance-level features of both domains can be aligned to have the same distribution conditional to the offset.

\subsection{Overall Loss}
Using labeled source domain data, the backbone and heads of FCOS are trained by minimizing object detection loss $\mathcal{L}_{det}$ consisting of object classification loss $\mathcal{L}_{det\mbox{-} cls}$ and bounding box regression loss $\mathcal{L}_{det\mbox{-} reg}$, as in \cite{tian2019fcos}:
\begin{equation}
    \begin{gathered}
        \mathcal{L}_{det}(x^S)= \mathcal{L}_{det\mbox{-} cls}+\mathcal{L}_{det\mbox{-} reg}.
    \end{gathered}
\label{eq:loss-det}
\end{equation}
In addition, we introduce $\mathcal{L}_{g}$ in (\ref{eq:loss-global}) to ensure that the overall features of both domains have the similar marginal distribution and $\mathcal{L}_{c}$ in (\ref{eq:cond_loss}) to allow the instance-level features to have the same conditional distribution to offsets, as follows:
\begin{equation}
    \begin{gathered}
        \mathcal{L}_{total} = \mathcal{L}_{det}(x^S) + \lambda_{g}\mathcal{L}_{g}(x^S, x^T) + \lambda_{c}(\mathcal{L}_{c}^{left} (x^S, x^T)+\mathcal{L}_{c}^{top} (x^S, x^T)).
    \end{gathered}
\label{eq:loss-total}
\end{equation}
where $\lambda_{g}$ and $\lambda_{c}$ are parameters balancing the loss components.

\section{Experiments}
\subsection{Datasets}

We conduct experiments on three scenarios: adaptation to adverse weather driving (Cityscapes to Foggy Cityscapes, i.e. CS $\rightarrow$ FoggyCS), adaptation from synthetic data to real data (Sim10k to Cityscapes, i.e. Sim10k $\rightarrow$ CS), and adaptation to a different camera modality (KITTI to Cityscapes, i.e. KITTI $\rightarrow$ CS).
\begin{itemize}[leftmargin=*]

\item \textbf{Cityscapes} \cite{cordts2016cityscapes} consists of clear city images under driving scenarios, summing to 2,975 and 500 images for training and validation, respectively. There are 8 categories, \textit{i.e.,} person, rider, car, truck, bus, train, motorcycle and bicycle.

\item \textbf{Foggy Cityscapes} \cite{sakaridis2018semantic} is a synthetic dataset that is rendered by adding fog to the  Cityscapes images. We use Cityscapes as the source, and Foggy Cityscapes as the target to simulate domain shift caused by the weather condition. 

\item \textbf{Sim10k} \cite{johnson2016driving} consists of 10,000 synthesized city images. For the adaptation scenario from synthetic data to real data, we set Sim10k as the source domain and Cityscapes as the target domain. Only \textit{car} class is considered.

\item \textbf{KITTI} \cite{geiger2012kitti} consisting of 7,481 images is a driving scenario dataset similar to Cityscapes, but there is a difference in camera modality. For adaptive scenarios to other camera modalities, we use KITTI as the source and Cityscapes as the target. Similar to Sim10k to Cityscapes, only \textit{car} category is used.  
\end{itemize}

\begin{table*}[t]
	\centering
	\caption{Results of Cityscapes $\rightarrow$ Foggy Cityscapes. EPM$^*$ denotes our re-implementation and GA$^\dagger$ is the result of the global alignment of Sec.\ref{subsec:global_alignment}. \textit{Source Only} is trained with only source domain without adaptation and \textit{Oracle} is trained with labeled target domain, providing the upper bound of UDA. }
	\label{tab:city2foggy}
	\resizebox{.97\textwidth}{!}{%
	\begin{tabular}{p{0.53\textwidth}P{0.17\textwidth}p{0.08\textwidth}p{0.08\textwidth}p{0.08\textwidth}p{0.08\textwidth}p{0.08\textwidth}p{0.08\textwidth}p{0.08\textwidth}p{0.08\textwidth}p{0.08\textwidth}}
    	\toprule
    	Method &Detector&person &rider &car &truck &bus &train &mbike &bicycle &$\text{mAP}^r_{0.5}$ \\
    	\midrule 
    	Source Only&\multirow{8}{*}{Faster-RCNN} &17.8 &23.6 &27.1 &11.9 &23.8 &9.1 &14.4 &22.8 &18.8\\
    	DAFaster \cite{chen2018domain} &&25.0 &31.0 &40.5 &22.1 &35.3 &20.2 &20.0 &27.1 &27.6\\
    	Selective DA \cite{zhu2019adapting}&&33.5 &38.0 &48.5 &26.5 &39.0 &23.3 &28.0 &33.6 &33.8\\
    	MAF \cite{he2019multi} &&28.2 &39.5 &43.9 &23.8 &39.9 &33.3 &29.2 &33.9 &34.0\\
    	SWDA\cite{saito2019strongweak}&&29.9 &42.3 &43.5 &24.5 &36.2 &32.6 &30.0 &35.3 &34.3\\
    	HTCN \cite{chen2020harmonizing} &&33.2 &47.5 &47.9 &31.6 &47.4 &40.9 &32.3 &37.1 &39.8\\
    	UMT \cite{deng2021unbiased} &&34.2 &48.8 &51.1 &30.8 &51.9&42.5 &33.9&38.2&41.2\\
    	MeGA-CDA \cite{vs2021mega} &&37.7&  49.0& 52.4& 25.4& 49.2& 46.9& 34.5& 39.0& 41.8\\
    	\midrule
    	Oracle && 37.2& 48.2& 52.7& 35.2& 52.2&  48.5& 35.3& 38.8& 43.5 \\
    	\midrule
    	Source Only&\multirow{9}{*}{FCOS} &30.2 &27.4 &34.2 &6.8 &18.0 &2.7 &14.4 &29.3 &20.4\\
    	EPM \cite{hsu2020epm} &&41.9 &38.7 &56.7 &22.6 &41.5 &26.8 &24.6 &35.5 &36.0\\
    	EPM$^*$ \cite{hsu2020epm}  &&44.9 &44.4 &60.6 &26.5 &45.5 &28.9 &30.6 &37.5 &39.9\\
    	SSAL \cite{munir2021ssal} &&45.1 &\textbf{47.4} &59.4 &24.5 &\textbf{50.0} &25.7 &26.0 &38.7 &39.6 \\
    	GA$^\dagger$ &&43.2&40.5&58.2&28.2&43.6&24.2&27.1&35.3&37.5\\
    	OADA (Offset-Left) &&46.2 &45.0 &62.2 &26.8 &49.0 &39.2 &33.1 &39.1 &42.6 \\
    	OADA (Offset-Top) &&45.9 &46.3 &61.8 &30.0 &48.2 &36.0 &34.2 &39.0 &42.7 \\
    	OADA (Offset-Left \& Top) &&\textbf{47.3} &45.6 &\textbf{62.8} &\textbf{30.7} &48.0 &\textbf{49.4} &\textbf{34.6} &\textbf{39.5} &\textbf{44.8} \\
    	OADA (Offset-Left \& Top + Self-Training) &&47.8 &46.5 &62.9 &32.1 &48.5 &50.9 &34.3 &39.8 &45.4 \\
    	\midrule
    	Oracle &&49.6 &47.5 &67.2 &31.3 &52.2 &42.1 &32.9 &41.7 &45.6 \\
    	\bottomrule
	\end{tabular}
	}
\end{table*}

\begin{table}[t]
\centering
\caption{Results of Sim10k, KITTI $\rightarrow$ Cityscapes. EPM$^*$ denotes the results of our re-implementations. GA$^\dagger$ is the result of the global alignment of Sec.\ref{subsec:global_alignment}.}
\label{tab:sim2city}
\resizebox{0.6\linewidth}{!}{
\begin{tabular}{p{0.5\textwidth}P{0.2\textwidth}p{0.1\textwidth}p{0.1\textwidth}}
\toprule
&&Sim10k&KITTI\\ \midrule
Method             & Detector                     & $\text{mAP}^r_{0.5}$ & $\text{mAP}^r_{0.5}$ \\ \midrule
Source Only        & \multirow{8}{*}{Faster-RCNN} & 34.3  & 30.2       \\
DAFaster \cite{zhu2019adapting}               &          & 38.9 & 38.5                \\
SWDA \cite{saito2019strongweak}             & & 40.1 & 37.9                 \\
MAF \cite{he2019multi}               &  & 41.1 & 41.0           \\
HTCN \cite{chen2020harmonizing} && 42.5 & - \\
Selective DA \cite{zhu2019adapting}             &          & 43.0 & 42.5                 \\
UMT \cite{deng2021unbiased} && 43.1 & - \\
MeGA-CDA  \cite{vs2021mega}         &  & 44.8 & 43.0            \\
\midrule
Oracle             & & 69.7    & 69.7             \\ \midrule
Source Only  & \multirow{8}{*}{FCOS}        & 40.4  & 44.2        \\
EPM \cite{hsu2020epm}               &        & 49.0 & 45.0              \\
EPM$^*$ \cite{hsu2020epm}              &   & 51.1 & 43.7      \\
SSAL\cite{munir2021ssal}      &      & 51.8 & 45.6 \\
GA$^\dagger$ &&49.7&43.1\\
OADA (Offset-Left) &  & 55.4 &  45.6            \\
OADA (Offset-Top) &  & 55.7 & 45.8  \\
OADA (Offset-Left \& Top) &  & \textbf{56.6} & \textbf{46.3}\\
OADA (Offset-Left \& Top + Self-Training) &  & 59.2 & 47.8\\
\midrule
Oracle             &                         & 72.7  & 72.7               \\ \bottomrule
\end{tabular}
}
\end{table}
\subsection{Implementation Details}
We use VGG-16 \cite{simonyan2015vgg} backbone and fully-convolutional head consisting of three branches of classification, regression and centerness following \cite{tian2019fcos}. For the domain discriminators, $D_g$ and $D_c$, fully-convolutional layers with the same structure as the head are used.
We initialize the backbone with the Image-Net pretrained model and reduce the overall domain gap using only object detection loss $\mathcal{L}_{det}$ and global alignment $\mathcal{L}_{g}$ at the beginning of training. Then, we train the model for 20k iterations with weight decay of 1e-4, initial learning rate of 0.02 for CS $\rightarrow$ FoggyCS, 0.01 for Sim10k $\rightarrow$ CS and 0.005 for KITTI $\rightarrow$ CS, respectively. We decay the learning rate at 12k and 18k iteration by the rate of one-tenth. During training, $\lambda_{g}$ and $\lambda_{c}$ are fixed as 0.01 and 0.1, respectively. We set the weight for the Gradient Reversal Layer (GRL) to 0.02 for global alignment and 0.2 for our conditional alignment. Also, we set the confidence threshold in (\ref{eq:mask}) as $\rho = 0.3$ for CS $\rightarrow$ FoggyCS and $\rho = 0.5$ for Sim10k, KITTI $\rightarrow$ Cityscapes to reduce the effects of incorrect predictions. We set $I$ to 6k which is the half of the first learning rate decay point and $\alpha_0$ in (\ref{eq:alpha-blending}) to 0.2. Input image is resized to 800 for shorter side, and 1333 or less for longer side following \cite{hsu2020epm,munir2021ssal,tian2019fcos}.

\subsection{Overall Performance}
In Table \ref{tab:city2foggy}, we compare the performance of our method (\textit{OADA}) with other existing methods in CS $\rightarrow$ FoggyCS setting. When \textit{EPM} \cite{hsu2020epm} based on FCOS is trained in the exact same setting as ours, the performance is 3.9\%p higher than what was reported. Conditional alignment on the offsets to the left and the top side of the bounding box improves performance by 22.2\%p and 22.3\%p respectively compared to \textit{Source Only}, by 5.7\%p and 6.0\%p compared to \text{GA$^\dagger$} which is only globally aligned and by 2.7\%p and 2.8\%p compared to \textit{EPM*} that we reimplemented. Since Foggy Cityscapes has 8 categories of objects and has various aspect ratios, when conditioning is performed on both the left and the top offsets, the additional performance gain is very larger by 2.1\%p or more, achieving the state-of-the-art regardless of the detector architecture. By initializing with the \textit{OADA (Offset-Left \& Top)} pre-trained model and training once more with self-training~\cite{liu2021unbiased}, we get a model almost similar to \textit{Oracle} lagging only by 0.2\%p. Detailed implementation of self-training is explained in the supplementary.

Table \ref{tab:sim2city} shows the adaptation results of Sim10k and KITTI $\rightarrow$ CS. In Sim10k $\rightarrow$ CS, conditional alignment using left and top offsets improves mAP by 15.0\%p and 15.3\%p, respectively, compared to \textit{Source-Only} and 4.3\%p and 4.6\%p over re-implemented \textit{EPM*}. Likewise, in KITTI $\rightarrow$ CS, our methods using left and top offsets improve the performance by 1.4\%p and 1.6\%p, respectively, over \textit{Source-Only} and 1.9\%p and 2.1\%p over \textit{EPM*}. In these two benchmarks, only the \textit{car} class is considered, so the gain of \text{OADA (Offset-Left \& Top)} is not as large as the multi-category setting, but there are still additional gain of 0.9\%p and 0.5\%p for Sim10k and KITTI. Our conditional aligning alone already achieves the state-of-the-art performance but greater performance can be obtained by employing the self-training as in OADA (Offset-Left \& Top + Self-Training).

\subsection{Analysis on Discriminability and Transferability}
Chen et. al \cite{chen2019discriminability} argued that aligning the feature distribution through adversarial alignment increases the transferability of features, but they did not take the discriminability into account, the ability to perform tasks well. They measure the discriminability via singular value decomposition (SVD) of the target domain feature maps and measure the transferability by estimating the corresponding angle between the feature spaces of the source and the target domain. Using their proposed metrics \cite{chen2019discriminability,yang2020mind}, we compare the discriminability and transferability of \textit{Ours (OADA)} with \textit{Source Only}, \textit{GA} of Sec. \ref{subsec:global_alignment}, \textit{EPM} \cite{hsu2020epm} and \textit{Oracle} models in CS$\rightarrow$Foggy CS. When $F^S=[f_1^S...f_{N_S}^S]$ and $F^T=[f_1^T...f_{N_T}^T]$ are feature matrix of the source and the target domain respectively, we apply SVD as follows:
\begin{equation}
\begin{gathered}
    F^S = U_S\Sigma_SV_S^\top,\ \ F^T = U_T\Sigma_TV_T^\top.
    \end{gathered}
\label{eq:svd}
\end{equation}

\begin{figure}[t]
  \centering
  \begin{minipage}{.67\linewidth}
    \centering
    \subcaptionbox{Singular Values}
      {\includegraphics[width=0.49\linewidth,height=80pt]{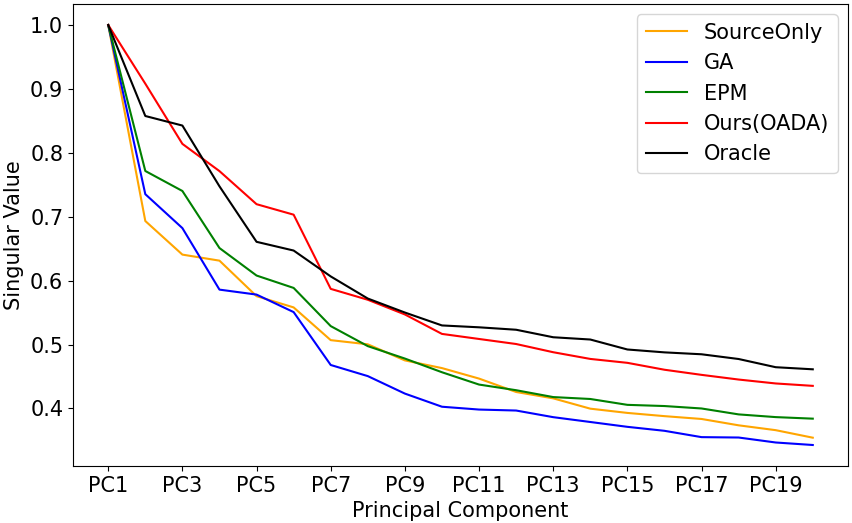}}
    \subcaptionbox{Corresponding Angles}
      {\includegraphics[width=0.49\linewidth,height=80pt]{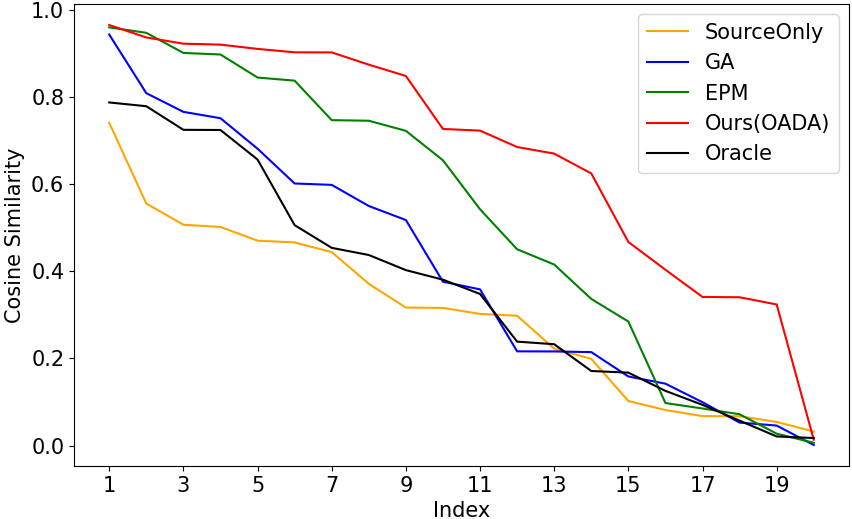}}
    \caption{Measures of discriminability and transferability}
    \label{fig:disc_trans}
  \end{minipage}\quad
  \begin{minipage}{.27\linewidth}
    \centering
    \includegraphics[width=.7\linewidth,height=97pt]{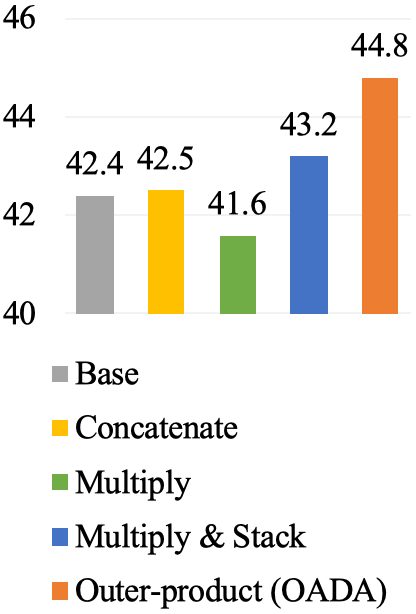}
    \caption{Conditioning}
    \label{fig:conditioning_strategies}
  \end{minipage}
\end{figure}
\noindent \textbf{Discriminability:} 
Fig. \ref{fig:disc_trans}a plots the top-20 greatest singular values of $F^T$ obtained from the five models. The singular values are sorted in descending order from left to right on the x-axis. These values are normalized so that the maximum singular value is 1. It can be seen that \textit{Ours} (red) has a similar decreasing ratio to the \textit{Oracle} model (black), while \textit{GA} and \textit{EPM} models have significantly larger one singular value and relatively much smaller other singular values than the largest one.
This means that informative signals corresponding to small singular values are greatly compromised in \textit{GA} and \textit{EPM}, where the entire features (GA) or the features close to the center of an object (EPM) are being aligned.
On the other hand, \textit{Ours} has a more gentle decreasing ratio, which shows that it maintains discriminability compared to other adversarial alignment methods.

\noindent \textbf{Transferability:} 
Transferability between the source and the target domain is measured through the similarity of each principal component of the two feature spaces \cite{chen2019discriminability}. Fig.~\ref{fig:disc_trans}b shows the cosine similarity of eigenvectors corresponding to the top 20 singular values in $F^S$ and $F^T$ obtained in descending order. While \textit{SourceOnly} and \textit{Oracle}, which do not perform feature alignment, have low corresponding angles of principal components between $F^S$ and $F^T$, overall correspondences increase in models that align feature spaces in an adversarial manner. Particularly, \textit{Ours} shows higher similarity between the source and the target domain feature space than \textit{GA} and \textit{EPM}. From this observation, it can be seen that aligning the features in an offset-aware manner is effective in increasing the transferability without harming the discriminability for object detection.

\subsection{Ablation Studies}
\label{subsec:ablation}
\begin{table}[t]
\centering
\caption{$\text{mAP}^r_{0.5}$ according to $\alpha_0$ of (\ref{eq:alpha-blending}) and number of bins in CS$\rightarrow$Foggy CS.}
\label{tab:alpha}
\resizebox{0.8\linewidth}{!}{
\begin{tabular}{{P{0.35\textwidth}|P{0.07\textwidth}P{0.07\textwidth}P{0.07\textwidth}P{0.07\textwidth}P{0.07\textwidth}|P{0.07\textwidth}P{0.07\textwidth}P{0.07\textwidth}P{0.07\textwidth}P{0.07\textwidth}}}
\toprule
\multirow{2}{*}{Model}& \multicolumn{5}{c|}{$\alpha_0$} & \multicolumn{5}{c}{$N_{bin}$}\\ 
\cline{2-6} \cline{7-11}
& 0.0      & 0.1      & 0.2 & 0.5 & 1.0 & 1 & 2 & 3 & 4 & 5   \\ \hline
OADA(Offset-Left \& Top)& 42.6    & 43.4    & \textbf{44.8}    & 43.3 & 42.6 & 42.4& 44.2 & \textbf{44.8} &44.3 & 41.4\\
\bottomrule
\end{tabular}
}
\end{table}

\textbf{$\alpha_0$ in (\ref{eq:alpha-blending}): }
Table \ref{tab:alpha} shows the effect of $\alpha_0$ that smoothes $\tilde{q}$ used for feature conditioning. When $\alpha_0=0$, conditioning is done using only $q$, which is the probability vector of an offset, without smoothing after the warm-up period. Since the softmax in (\ref{eq:reg-bin}) makes $q$ almost one-hot vector in most cases, the values of certain rows of conditioned feature $g$ are almost zero. This strongly constrains the dimension of the conditioned features. On the other hand, when $\alpha_0=1$, we outer-product the feature with an uniform vector $\frac{1}{N_{bin}}\mathbbm{1}$ without offeset-aware conditioning throughout training. In this case, the feature dimension given to the discriminator is the same, but performance is greatly degraded because there is no conditioning according to the offset. $\alpha_0=0.2$, in which relatively strongly constrained features are used in the discriminator, is most appropriate.

\noindent
\textbf{Number of bins:}
\label{subsection:ablation_bin_number}
Table \ref{tab:alpha} also shows ablations results of the number of bins, $N_{bin}$.
The reference value $m_i$ for the $i$-th bin is set according to the size of objects for which the feature of each level is responsible. The detailed values are in the supplementary.
When $N_{bin}=1$, there is no conditioning on offsets and only the mask $M_{obj}$ is used, nevertheless it is more effective than \textit{GA} and \textit{EPM}.
$N_{bin}=3$ is sufficiently effective in offset conditioning, improving $\text{mAP}^r_{0.5}$ by 2.4\%p.
When the number of bins is increased excessively, the constraint on the subspace of the conditioned feature becomes too strong, resulting in performance deterioration. 

\noindent
\textbf{Conditioning strategies:} We compare various conditioning strategies in Fig. \ref{fig:conditioning_strategies}. When the feature and offset values are simply concatenated (\textit{Concatenate}) and the feature is multiplied by offset values (\textit{Multiply}), there is a very slight performance improvement of 0.1\%p or performance degradation of -0.8\%p compared to the case when only the unconditioned feature is used (\textit{Base}). In order to compare the method of conditioning while increasing the dimension of features to the same as \textit{OADA}, we also experiment with the case, \textit{Multiply\&Stack}, where original features, features multiplied by top (left) offsets and features multiplied by bottom (right) offsets are stacked. In this case, there is a significant performance improvement, but it is still far behind our \textit{OADA}, which means that it is much more effective to convert the offset value into a probability vector and outer-product it with the feature for conditioning.

\begin{table}[t]
\centering
\caption{Comparison of $\text{mAP}^r_{0.5}$ according to the confidence threshold ($\rho$)}
\label{tab:confidence}
\resizebox{0.7\linewidth}{!}{
\begin{tabular}{{P{0.2\textwidth}P{0.45\textwidth}P{0.08\textwidth}P{0.08\textwidth}P{0.08\textwidth}P{0.08\textwidth}}}
\toprule
\multirow{2}{*}{Model}&\multirow{2}{*}{Datasets} & \multicolumn{4}{c}{Confidence Threshold ($\rho$)} \\ \cline{3-6} 
                                         && 0.0      & 0.3      & 0.5      & 0.7     \\ \hline
\multirow{3}{*}{OADA(Offset-Top)}&Cityscapes $\rightarrow$ Foggy Cityscapes & 37.0    & \textbf{42.7}    & 41.7    & 37.3   \\
&Sim10k $\rightarrow$ Cityscapes           & 48.5    & 53.6    & \textbf{55.7}    & 53.6   \\
&KITTI $\rightarrow$ Cityscapes            & 43.8    & 44.8    & \textbf{45.8}    & 44.0\\
\bottomrule
\end{tabular}
}
\end{table}

\noindent
\textbf{Confidence thresholds:}
\label{subsection:albation_confidence_th}
In (\ref{eq:mask}), the confidence threshold, $\rho$, is used to generate a mask which activates spatial locations with high objectness and accurate regression values for the target domain.
Table \ref{tab:confidence} shows ablation results for $\rho$ when conditioning is performed only on the top offsets. In CS $\rightarrow$ Foggy CS, the performance is best when $\rho = 0.3$ and in Sim10k, KITTI $\rightarrow$ CS when $\rho = 0.5$. We conjecture that it is due to the difference of the domain gap between the source and the target domain in each scenario. Referring to Fig.~\ref{fig:Conf-CS}, in the case of CS $\rightarrow$ Foggy CS, when the confidence is more than 0.3, the precision is already more than 0.9 and the difference between GT and regression value is less than 5.
However, in the case of Sim10k $\rightarrow$ CS, the confidence must be at least 0.5 to obtain a similar level of precision and difference as shown in Fig.~\ref{fig:Conf-Sim10k}. This shows that $\rho$ must be adjusted with respect to the domain gap.


\section{Conclusions}
In this paper, we propose an Offset-Aware Domain Adaptive object detection method which conditionally aligns the feature distribution according to the offsets. Our method improves both discriminability and transferability by addressing negative transfer considering the modality of the feature distribution of an anchor-free one-stage detector. On various benchmarks, ours also achieves the state-of-the-art performance by significantly outperforming existing methods.\\
\noindent  \textbf{Acknowledgments} 
This work was supported by the National Research Foundation of Korea (NRF) grant (2021R1A2C3006659) and IITP grant (NO.2021-0-01343, Artificial Intelligence Graduate School Program - Seoul National University), both funded by the Korea government (MSIT). It was also supported by SNUAILAB.
\clearpage
%
%
\bibliographystyle{splncs04}
\bibliography{egbib}
\end{document}


\pagestyle{headings}
\mainmatter
\def\ECCVSubNumber{3958}  

\title{Supplementary Materials for \\Unsupervised Domain Adaptation for One-Stage \\Object Detector using Offsets to Bounding Box} 


\author{}
\institute{}
\maketitle

\appendix
\section{Binning Strategies}

\subsection{The ratio of features assigned to each bin}
\begin{figure*}[t]
\centering
  \begin{subfigure}[t]{6cm}
    \includegraphics[width=\linewidth,height=4cm]{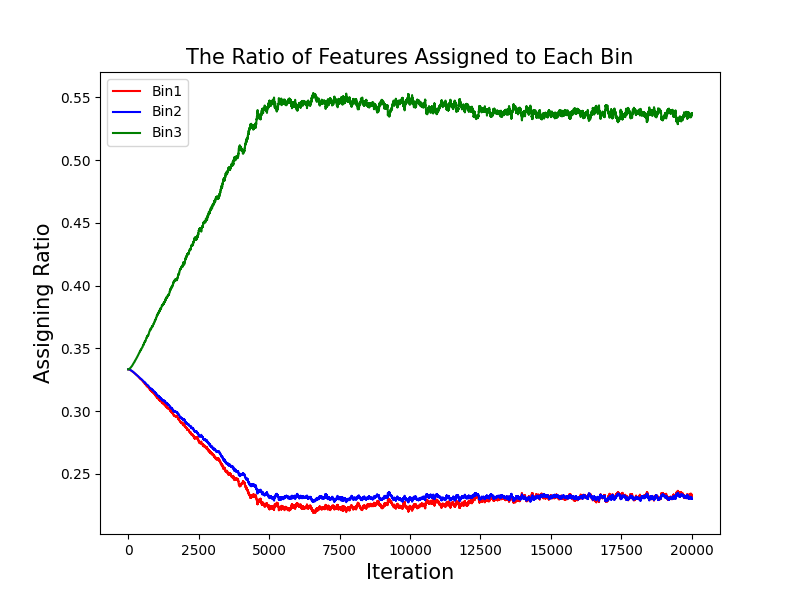}
    \caption[width=\linewidth]{Based on the left offsets in $F_3$}
    \label{fig:F3_left}
  \end{subfigure}
  \centering
  \begin{subfigure}[t]{6cm}
    \includegraphics[width=\linewidth,height=4cm]{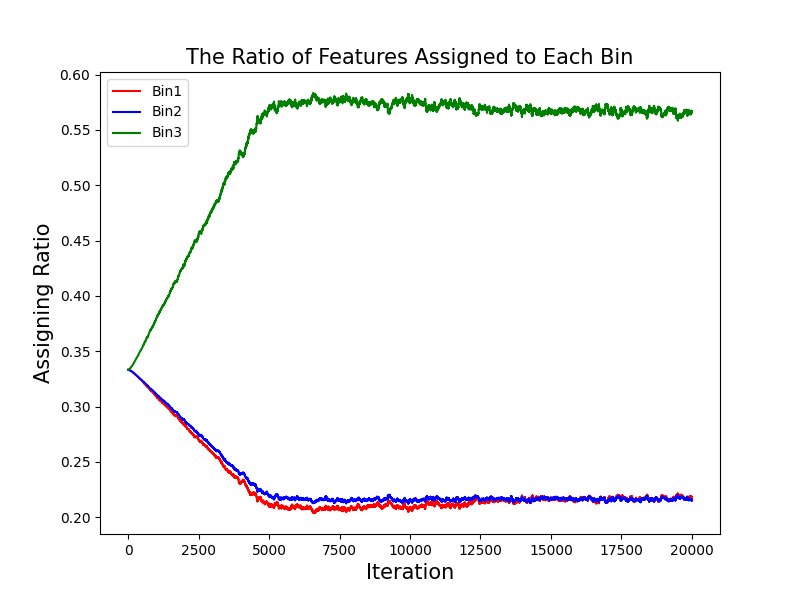}
    \caption{Based on the top offsets in $F_3$}
    \label{fig:F3_top}
  \end{subfigure}
  \centering
  \begin{subfigure}[t]{6cm}
    \includegraphics[width=\linewidth,height=4cm]{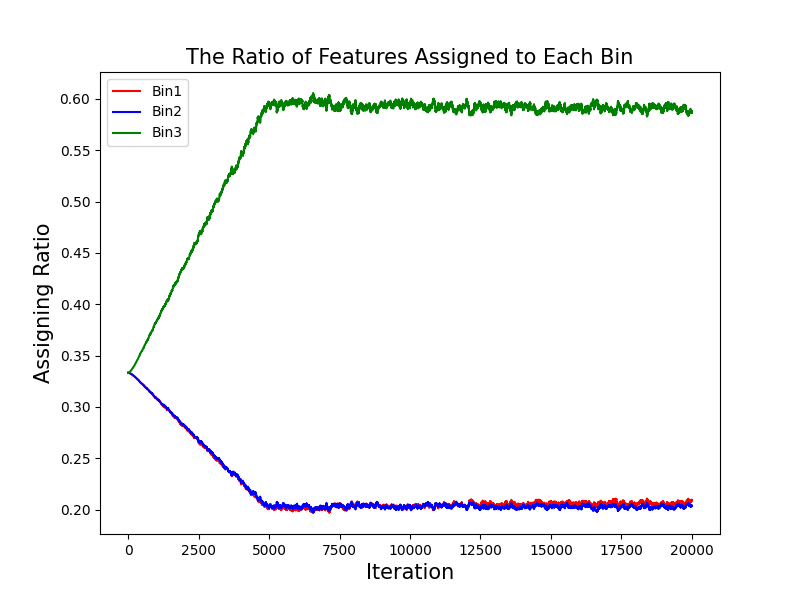}
    \caption{Based on the left offsets in $F_4$}
    \label{fig:F4_left}
  \end{subfigure}
\centering
  \begin{subfigure}[t]{6cm}
    \includegraphics[width=\linewidth,height=4cm]{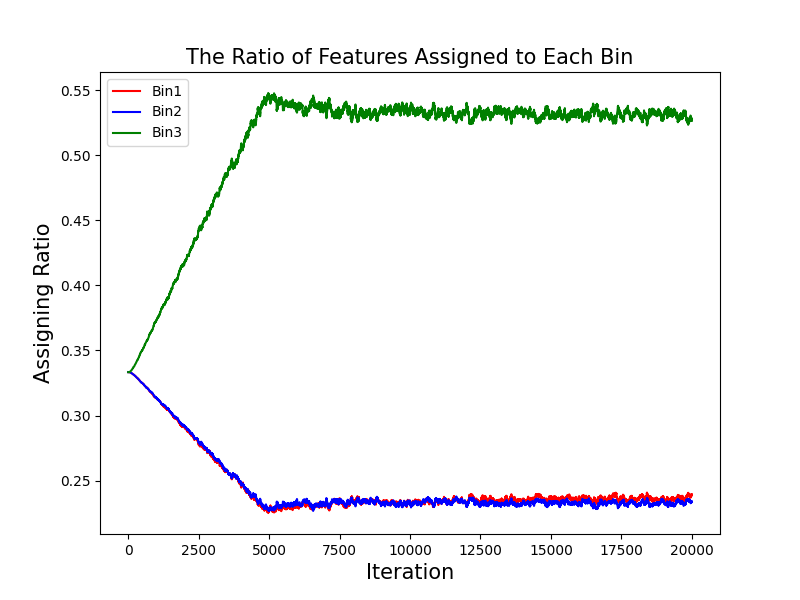}
    \caption{Based on the top offsets in $F_4$}
    \label{fig:F4_top}
  \end{subfigure}
\centering
  \begin{subfigure}[t]{6cm}
    \includegraphics[width=\linewidth,height=4cm]{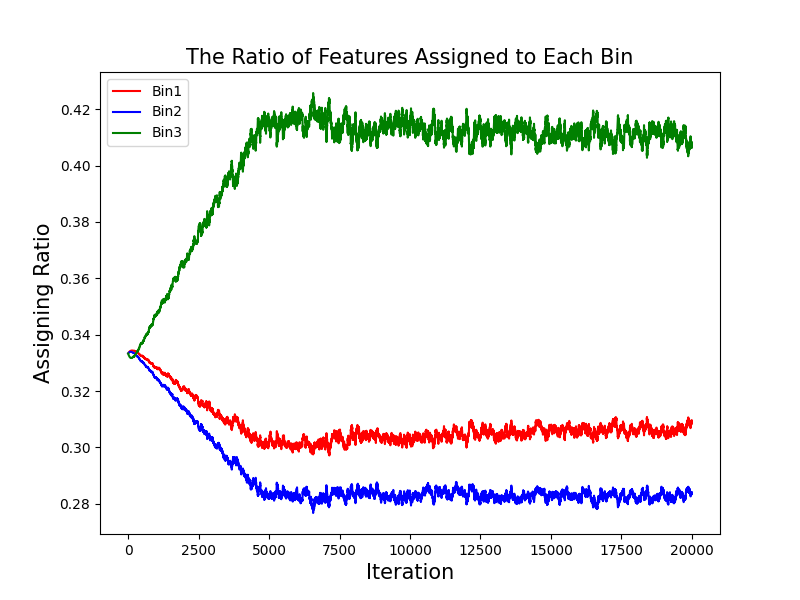}
    \caption{Based on the left offsets in $F_5$}
    \label{fig:F5_left}
  \end{subfigure}
    \centering
    \begin{subfigure}[t]{6cm}
    \includegraphics[width=\linewidth,height=4cm]{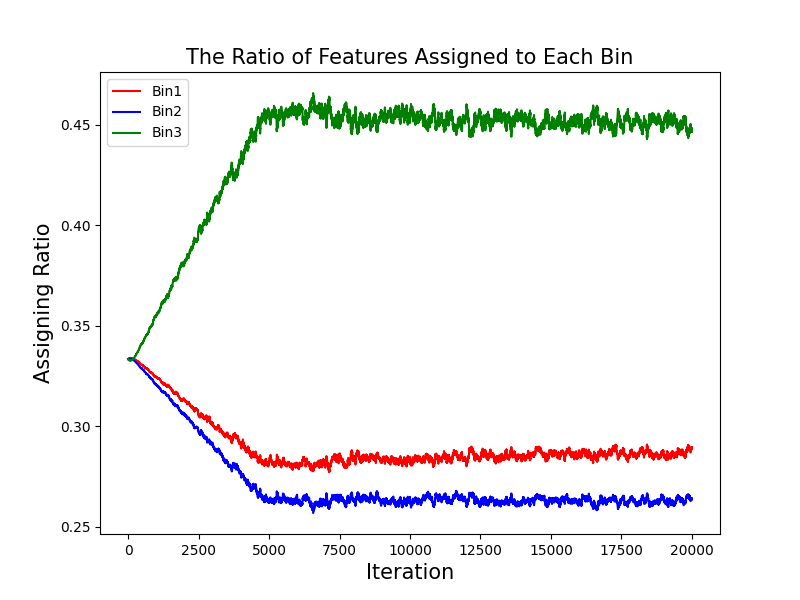}
    \caption{Based on the top offsets in $F_5$}
    \label{fig:F5_top}
  \end{subfigure}
  \centering
    \begin{subfigure}[t]{6cm}
    \includegraphics[width=\linewidth,height=4cm]{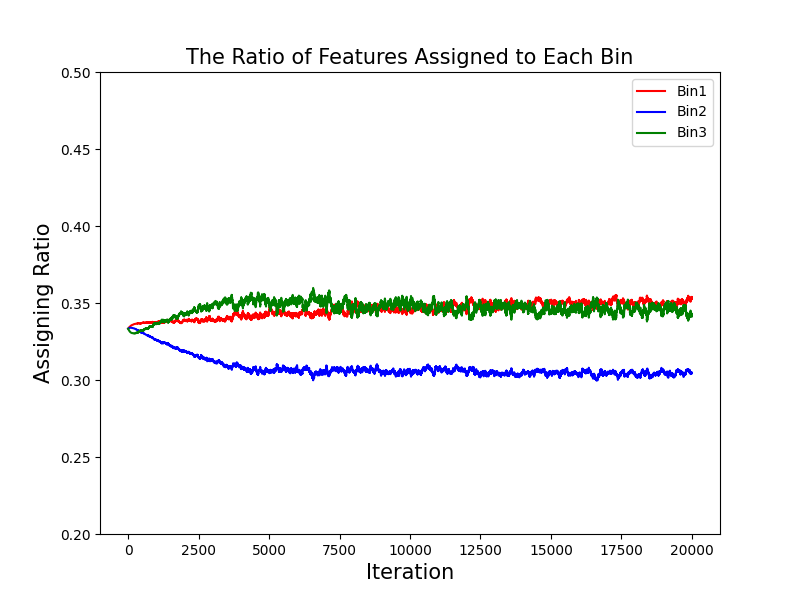}
    \caption{Based on the left offsets in $F_6$}
    \label{fig:F6_left}
  \end{subfigure}
  \centering
      \begin{subfigure}[t]{6cm}
    \includegraphics[width=\linewidth,height=4cm]{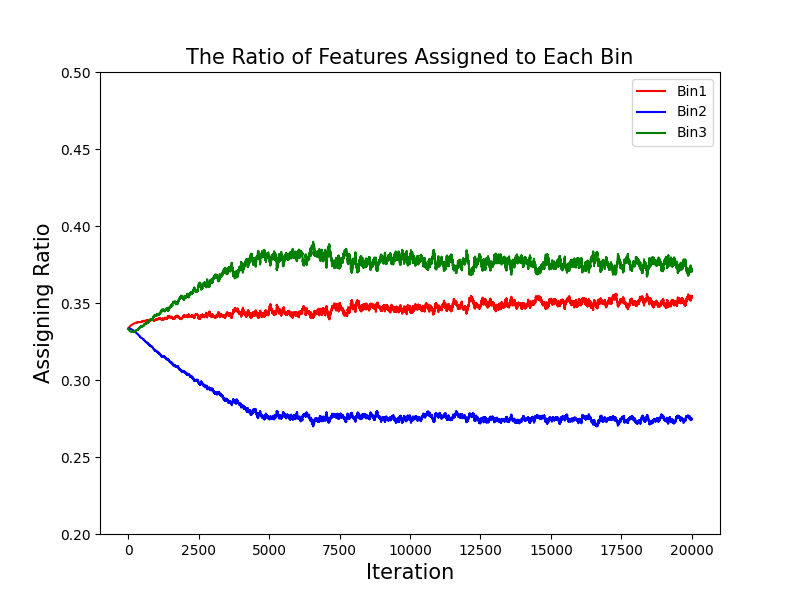}
    \caption{Based on the top offsets in $F_6$}
    \label{fig:F6_top}
  \end{subfigure}
  \caption{
  The ratio of feature allocated to each bin at different feature level as the iteration progresses during training.
  }
  \label{fig:Binning_Ratio}
 \vspace{-4mm}
\end{figure*}

In Sec. 3.4 of the main paper, we set $m_i$ of each bin to $(m_1,m_2,m_3)=(lv-\frac{1}{2},lv+\frac{1}{2},lv+\frac{3}{2})$ for each $lv$-th level feature to convert the offset value into a categorical probability vector. 
Fig. \ref{fig:Binning_Ratio} shows the average of the probabilities that the left $l$ and top $t$ offset values belong to each bin as the training iteration progresses in the CS $\rightarrow$ FoggyCS setting at different feature level.
Initially, the probability of the offset value belonging to each bin begins as a uniform distribution so that $(\frac{1}{3}, \frac{1}{3}, \frac{1}{3})$, and it gradually converges to the average of $\tilde{q}$, the probability vector of the offset values, during the warm-up period ($I$ iterations). 
Since $m_i$ is not finely set to make the the ratio of the offset values corresponding to each bin equal, in the case of $F_3$ and $F_4$, the ratio belonging to the last bin is somewhat high. 
However, setting the value of $m_i$ as mentioned in the main paper is sufficiently effective. 
This is because it is not important to ensure that the proportion of features belonging to each bin is equal, but important to similarly match the distribution between the features of the two domains belonging to the same bin.

\subsection{Setting of the $m_i$ values according to the number of bins}
Table 3 of Sec. 4.5 in the main paper compares the effect of conditioning the feature more strongly or loosely by changing the number of bins. Table \ref{tab:m_values} specifies $m_i$ values for each $N_{bin}$ setting at different feature level. We set $m_i$ values so that more than a certain ratio of features can be assigned to each bin.

\begin{table}[t]
\centering
\caption{$m_i$ values set according to $N_{bin}$ and different feature level.}
\label{tab:m_values}
\resizebox{.95\textwidth}{!}{%
\begin{tabular}{c|c|c|c|c}
\toprule
& $N_{bin}=2$   & $N_{bin}=3$      & $N_{bin}=4$        & $N_{bin}=5$               \\ \hline
Feature Level  & $(m_1, m_2)$   & $(m_1, m_2, m_3)$    & $(m_1, m_2, m_3, m_4)$     & $(m_1, m_2, m_3, m_4, m_5)$      \\ \hline
$F_3$             & (3.5, 4.5) & (2.5, 3.5, 4.5) & (2.5, 3.5, 4.5, 5.5) & (2.5, 3.5, 4.5, 5.5, 6.5) \\
$F_4$             & (4.5, 5.5) & (3.5, 4.5, 5.5) & (3.5, 4.5, 5.5, 6.5) & (3.5, 4.5, 5.5, 6.5, 7.5) \\
$F_5$             & (5.5, 6.5) & (4.5, 5.5, 6.5) & (4.5, 5.5, 6.5, 7.5) & (4.5, 5.5, 6.5, 7.5, 8.5) \\
$F_6$             & (6.5, 7.5) & (5.5, 6.5, 7.5) & (4.5, 5.5, 6.5, 7.5) & (4.5, 5.5, 6.5, 7.5, 8.5) \\
$F_7$             & (7.5, 8.5) & (6.5, 7.5, 8.5) & (5.5, 6.5, 7.5, 8.5) & (5.5, 6.5, 7.5, 8.5, 9.5) \\
\bottomrule
\end{tabular}
}
\end{table}

\section{Training Details for the Self-Training}

We basically follow the details of the method proposed in Liu et. al \cite{liu2021unbiased}. Even though \cite{liu2021unbiased} is a work that tackles semi-supervised object detection, its proposed method is applicable to unsupervised domain adaptation as well since the main idea is about how to handle the unlabeled data. It suggests how to train the object detector with the unlabeled data in a self-training manner using the unbiased teacher network. Only difference in the setting of ours from \cite{liu2021unbiased} is that in \cite{liu2021unbiased}, the labeled and the unlabeled data are from the same distribution, but in our case of unsupervised domain adaptation, the unlabeled data are from different distribution of the labeled data. 
The idea of \cite{liu2021unbiased} is to utilize the unbiased teacher network to produce the pseudo labels for the unlabeled data. The teacher object detector has the same architecture as the student but its parameters are not optimized by gradient descent but are updated by exponential moving average of the student network, as suggested in Tarvainen et. al \cite{tarvainen2018mean}. The parameters of the teacher detector are updated as follows:
\begin{equation} 
    \theta_t = \alpha\theta_t + (1-\alpha)\theta_s
    \label{eq:ema}
\end{equation}
where $\theta_t$ and $\theta_s$ represent the parameters of the teacher and the student network, respectively. $\alpha$ is the EMA rate which decides the percentage of the parameters of teacher network in the previous time step to be applied to the updated teacher parameters. The larger the $\alpha$, the slower the teacher network progresses. We empirically find that $\alpha=0.9999$ works the best in our setting of unsupervised domain adaptation. Also, the EMA interval indicates the number of iterations between EMA updates. EMA interval is set differently for each benchmark, for CS $\rightarrow$ FoggyCS, it is set to 10, while for KITTI $\rightarrow$ CS and Sim10k $\rightarrow$ CS, it is set to 5 and 1. The smaller the value, the more frequent the EMA updates are.

In \cite{liu2021unbiased}, the predictions of the teacher network are used as the pseudo labels of the target domain to train the student network. \cite{liu2021unbiased} feeds weakly augmented unlabeled data into the teacher network and strongly augmented unlabeled data are fed into the student, by differentiating inputs to the two networks, resulting in knowledge gaps between the predictions of the teacher and the student networks. The student tries to narrow this gap by training unlabeled data with pseudo labels generated by the teacher network. However, in our case, we find that applying weak augmentation to the inputs of the teacher network unnecessary and using the original target inputs is effective. For example, when given a target image, $x^T$, the teacher object detector, $f_t$, predicts a set of bounding boxes.
\begin{equation} 
    \mathbb{B}_t^T = ({\hat{y}_i^T, \hat{b}_i^T)}_{i=1}^{k} = f_t(x^T)
\end{equation} 
$\hat{y}_i \in \mathbb{R}^C$ where $0\leq \hat{y}_{i,c} \leq 1$ and $\hat{b}_i \in \mathbb{R}^{4}$ indicates the classification confidence and the predicted box coordinates $(l, t, r, b)$, respectively. $k$ is the number of bounding boxes predicted for a given target domain image, $x^T$, in the teacher detector . Note that $T$ in the superscript refers to the `target domain' and the $t$ in the subscript refers to the `teacher detector'. Then we threshold the predicted boxes from the teacher detector using the confidence score. Specifically, we set a threshold $\delta$ and eliminate bounding boxes with the confidence score less then or equal to $\delta$. Therefore, $\mathbb{B'}_t^T = \{(\hat{y}_i^T, \hat{b}_i^T)|\max_{c}(\hat{y}_{i,c})>\delta\} \subset \mathbb{B}_t^T$. $\delta$ is set as 0.5 empirically in all of our experiments.
Finally, we use this thresholded bounding boxes from the teacher network as the pseudo labels of the target domain to train the student detector. As mentioned earlier, strongly augmented inputs are fed into the student detector, where the same strong augmentation strategy as described in \cite{liu2021unbiased} is used. The overall loss function to train the student network, $f_s$, for both the source and the target domain is as follows:
\begin{equation} 
    \begin{gathered}
        \mathcal{L}_{student} = \mathcal{L}_{det}(x^S, (y^S,b^S)) + \lambda_{self}\mathcal{L}_{det}(\mathcal{A}(x^T), \mathbb{B'}_t^T)    
    \end{gathered}
\end{equation} 
where $\mathcal{A}$ refers to the strong augmentation and $\lambda_{self}$ is the weight on the self-training loss for the target domain. Here, we use $\lambda_{self}=2$ since it shows the best results. While $f_s$ is trained by above loss function, on the other hand, the teacher network is updated via EMA of (\ref{eq:ema}) as explained earlier. 

In all of our self-training experiments, the student and the teacher detectors are initialized by the detector that is pre-trained with our proposed OADA (Offset-Left \& Top), so that the $f_t$ is able to produce reasonably correct pseudo labels from the beginning of the training. We report the performance of the teacher detector since it shows highly improved performance due to the temporal ensemble effect.
We only applied the proposed method of \cite{liu2021unbiased} to our problem, so there is no contribution in terms of novelty, but we consider this experiment as an important study result because we applied the proposed method to an anchor-free one-stage detector, FCOS for the first time. Considering that the method was originally proposed for Faster R-CNN, a two-stage detector, this experimental results further prove the applicability of the method to other detector architectures and the unsupervised domain adaptation setting.

\section{Qualitative Results}
\begin{figure*}[t]
    \centering
    \includegraphics[width=.9\linewidth]{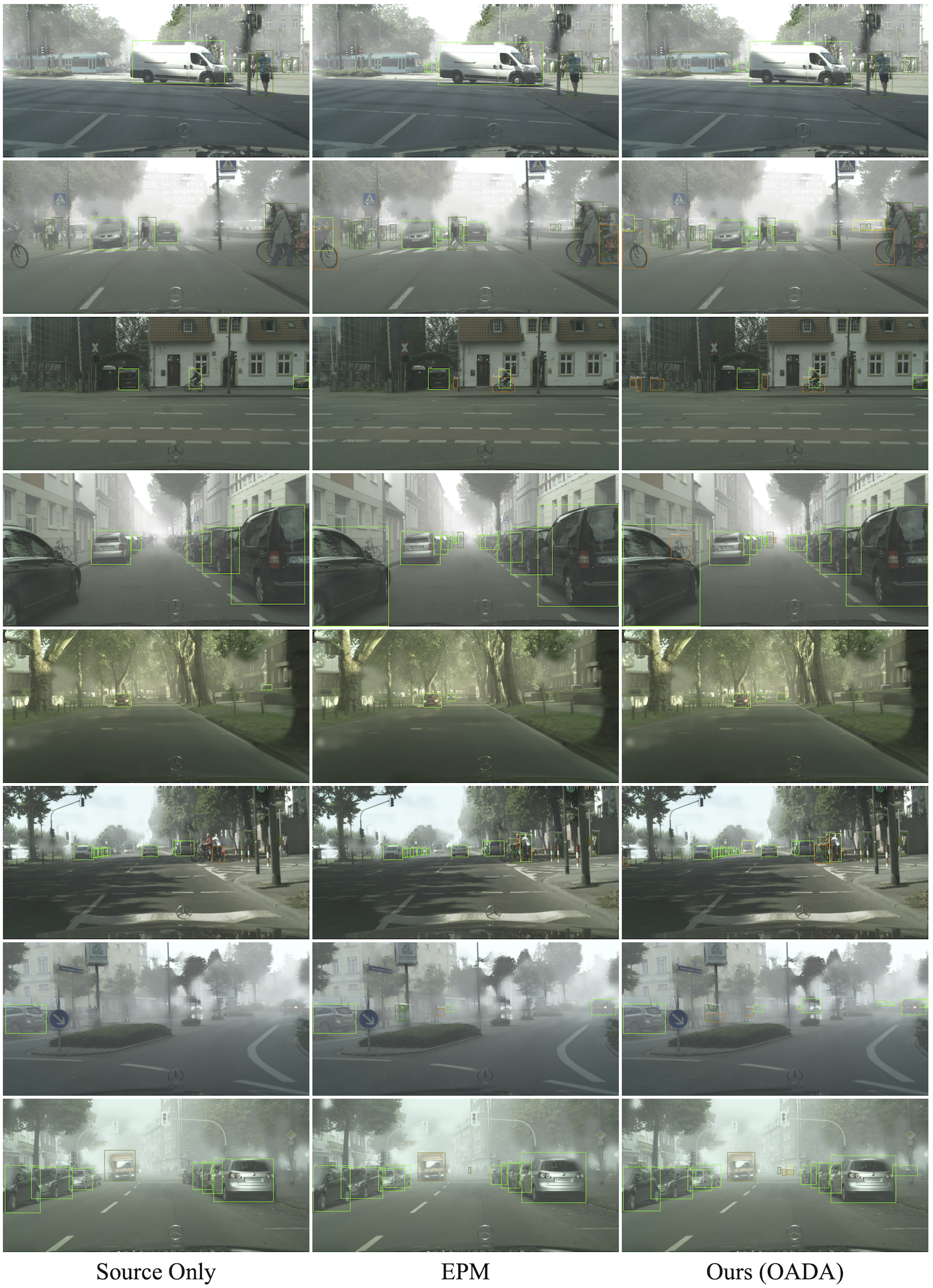}
    \caption{Qualitative results of \textit{SourceOnly}, \textit{EPM}\cite{hsu2020epm} and \textit{OADA} in CS $\rightarrow$ FoggyCS.}
    \label{fig:qualitative_results_CS2Foggy}
    \vspace{-5mm}
\end{figure*}

\begin{figure*}[t]
    \centering
    \includegraphics[width=.9\linewidth]{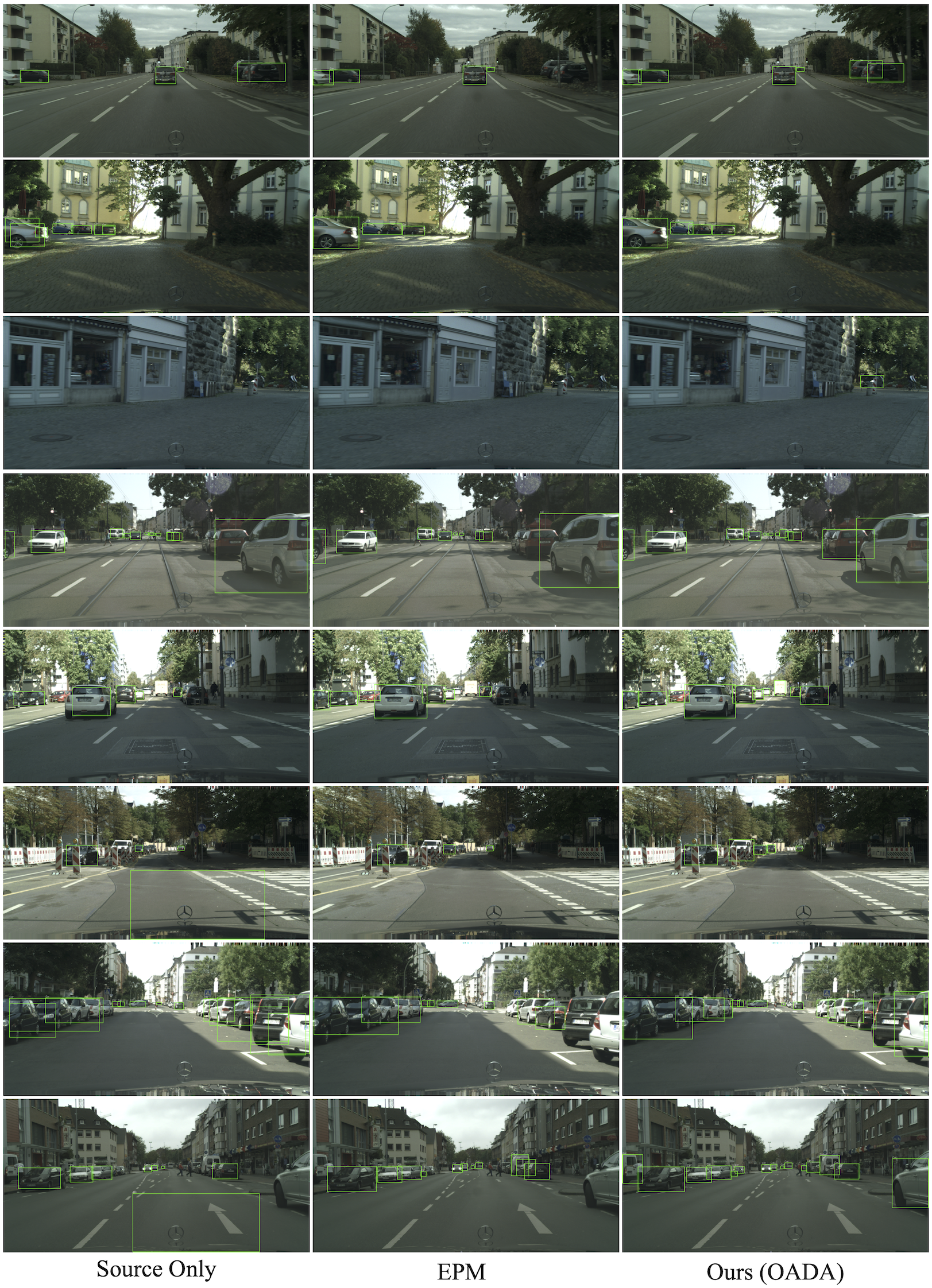}
    \caption{Qualitative results of \textit{SourceOnly}, \textit{EPM}\cite{hsu2020epm} and \textit{OADA} in Sim10k $\rightarrow$ CS.}
    \label{fig:qualitative_results_Sim10k2CS}
    \vspace{-5mm}
\end{figure*}

\begin{figure*}[t]
    \centering
    \includegraphics[width=.9\linewidth]{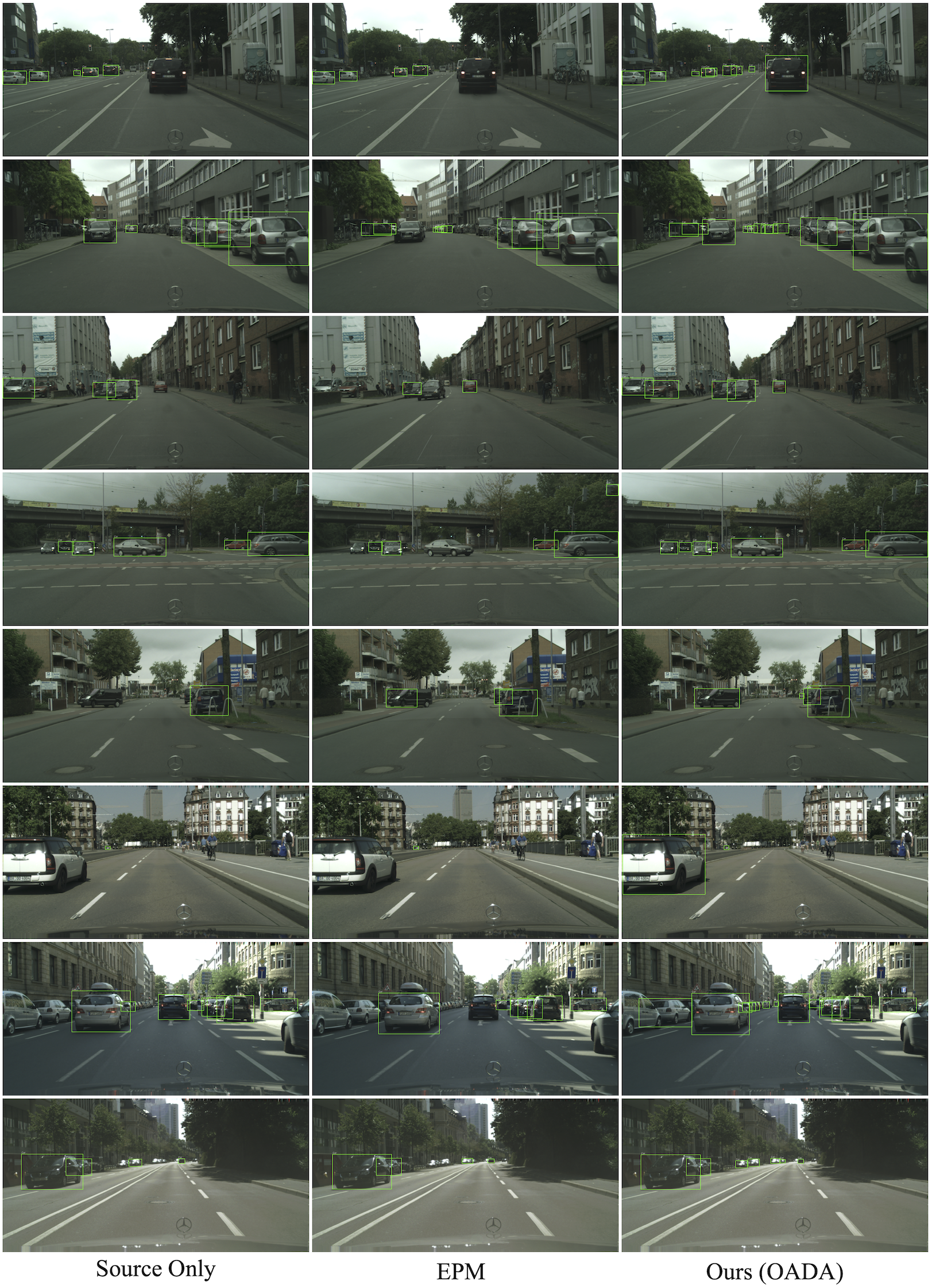}
    \caption{Qualitative results of \textit{SourceOnly}, \textit{EPM}\cite{hsu2020epm} and \textit{OADA} in KITTI $\rightarrow$ CS.}
    \label{fig:qualitative_results_KITTI2CS}
    \vspace{-5mm}
\end{figure*}

Fig.\ref{fig:qualitative_results_CS2Foggy}, \ref{fig:qualitative_results_Sim10k2CS}, and \ref{fig:qualitative_results_KITTI2CS} are the qualitative results of the \textit{SourceOnly}, \textit{EPM}\cite{hsu2020epm} and \textit{OADA} methods in CS $\rightarrow$ FoggyCS, Sim10k $\rightarrow$ CS, and KITTI $\rightarrow$ CS benchmarksets, respectively. As shown in the figure, \textit{OADA} is able to detect distant objects in the center better than baseline methods.

\clearpage
%
%
\bibliographystyle{splncs04}
\bibliography{egbib}